# GiesKaNe: Bridging Past and Present in Grammatical Theory and Practical Application

Volker Emmrich[1]

**Abstract** This article explores the requirements for corpus compilation within the GiesKaNe project (University of Giessen and Kassel, Syntactic Basic Structures of New High German). The project is defined by three central characteristics: it is a reference corpus, a historical corpus, and a syntactically deeply annotated treebank. As a historical corpus, GiesKaNe aims to establish connections with both historical and contemporary corpora, ensuring its relevance across temporal and linguistic contexts. The compilation process strikes the balance between innovation and adherence to standards, addressing both internal project goals and the broader interests of the research community. The methodological complexity of such a project is managed through a complementary interplay of human expertise and machine-assisted processes. The article discusses foundational topics such as tokenization, normalization, sentence definition, tagging, parsing, and inter-annotator agreement, alongside advanced considerations. These include comparisons between grammatical models, annotation schemas, and established de facto annotation standards as well as the integration of human and machine collaboration. Notably, a novel method for machine-assisted classification of texts along the continuum of conceptual orality and literacy is proposed, offering new perspectives on text selection. Furthermore, the article introduces an approach to deriving de facto standard annotations from existing ones, mediating between standardization and innovation. In the course of describing the workflow the article demonstrates that even ambitious projects like GiesKaNe can be effectively implemented using existing research infrastructure, requiring no specialized annotation tools. Instead, it is shown that the workflow can be based on the strategic use of a simple spreadsheet and integrates the capabilities of the existing infrastructure.

## Table of Contents

---

[1] Volker Emmrich; University of Gießen, Germany; volker.emmrich@germanistik.uni-giessen.de.

# 1. GiesKaNe: reference corpus, historical corpus, treebank

The GiesKaNe corpus (= Gies[sen]Ka[ssel]Ne[uhochdeutsch])[2] has been compiled since 2016 as part of the long-term project "Syntaktische Grundstrukturen des Neuhochdeutschen" (Engl.: Syntactic basic structures of New High German) funded by the German Research Foundation (DFG). The process of its compilation is presented here using the 3 aspects that characterize the corpus: it is a reference corpus, a treebank and a historical corpus.

The predicate 'reference' in reference corpus is used more often than it is defined, although there seems to be a core agreement about central criteria. As a reference corpus GiesKaNe is intended to reflect the language as comprehensively as possible so that it can also be used as a control corpus when evaluating special corpora (Lemnitzer & Zinsmeister, 2015: 141)[3]. So compared to special corpora, which are created with a specific research interest in mind, the community idea is central here from the beginning (Emmrich & Hennig, 2022: 201) and it constantly accompanies the process of building the corpus. Sinclair (1996) refers 'comprehensive' to the effect – "designed to provide comprehensive information about a language" – and Barth and Schnell (2021: 34) describe reference corpora as "a standard reference for corpus-based research on a given language". This statement ultimately makes a point that becomes particularly clear in Leech's statement (2002): "[H]opefully it will be treated by its user community as some kind of 'standard'." It is all about requirements that shape the process of compilation, right down to circumstances that ultimately only affect reception and over which the editors of the corpus have no influence. What is essential here is the requirement to be large or 'comprehensive' enough to reflect the relevant variation of the language (e.g., formality, speech, preparedness and broad subject-matter) and to be able to serve as a basis for grammars, dictionaries and thesauri (Sinclair, 1996). For GiesKaNe, grammar is the point of reference. Sinclair points out that "a rough kind of representativeness is achieved by ensuring that a large quantity of text exemplifies each of these parameters", but 'large quantities' are not a viable strategy for a treebank of New High German and the process of text selection becomes all the more crucial. I will go into that briefly before I also include the property of GiesKaNe being a treebank.

Fišer and Lenardič (2020), who make a list of around 30 available reference corpora for 21 languages for the CLARIN infrastructure, refer to 'comprehensive' as the genres taken into account. Dittmann et al. (2012: 341) structure this requirement when they provide appropriate variation of the texts via an even compilation of the 4 'text types': (1) fiction, (2) non-fiction/science, (3) newspapers/journals, and (4 ) functional texts (such as recipes, protocols, or manuals) which are also central to the selection of texts in the 'Deutsches Textarchive' (DTA[4]). The DTA is the most extensive corpus for New High German[5] and Geyken and Gloning (2015: 167) describe it more as an archive than a reference corpus, while highlighting its potential as the basis of a reference corpus (2015: 175): "As stated above, there may be more than one reference corpus at hand depending on the research purpose." What I take away from this is that even reference corpora must always be seen as such relative to a research question – including or especially the constantly growing[6] ones like DeReKo[7], which are better described as an "Archive of General Reference Corpora of Contemporary Written German" (Kupietz et al., 2010), but of course also the historical ones. So reference corpora are not absolute. It is about what they can offer. GiesKaNe has to be seen in connection with the 'Deutsch Diachron Digital' (DDD)[8] network, in which the following reference corpora were developed: Old German (ReA)[9], Middle High German (ReM)[10], Early New High German (ReF)[11], Middle Low German/Lower Rhine ('Niederrheinisch') (ReN)[12] and 'Deutsche Inschriften' (Engl: German inscriptions)[13] (Dipper & Kwekkeboom, 2018). GiesKaNe closes the gap between the reference corpus of Early New High German and contemporary German. While the question of text selection for Old German hardly arises, the criteria of time period, area and genre are decisive for the other historical reference corpora. And GiesKaNe has to build on this as a supplementary reference corpus. As Fig. 1 shows,

---

GiesKaNe does not use newspaper texts as the DTA and its contemporary language counterpart DWDS[14], but instead uses texts from everyday life. The reference corpus of Early New High German, in turn, covers various linguistic landscapes and seven time periods between 1350 and 1700. Bridging the gap to contemporary German means doing justice to both language levels and a sensible positioning alongside the extensive DTA.

| | 17th century | 18th century | 19th century |
|---|---|---|---|
| everyday texts | 72.000 words[15] (= 6 texts) | 72.000 words (= 6 texts) | 72.000 words (= 6 texts) |
| science | … | … | … |
| utility literature | … | … | … |
| fiction | … | … | … |
| **TOTAL** | 288.000 words | 288.000 words | 288.000 words |

Fig. 1: planned overall structure of GiesKaNe with a focus on centuries and text types

The concept of 'immediacy and distance' ('Nähe und Distanz') by Koch and Oesterreicher (1985) is crucial for corpus compilation here, but corresponding categories are also valuable for the metadata and later analyses. Koch & Oesterreicher's model is theory-driven, but its implementation (cf. Emmrich, 2024) is comparable to Biber's Dimension 1 (1988), whose reception in German studies is not even remotely comparable to that in English studies (cf. Raible 2019; Schaeffer, 2021). However, the implementation of 'Nähe und Distanz' (Emmrich, 2024) is a controlled, quasi-tailored dimension between conceptual orality and literacy. Considering the theory-driven nature of the model of Koch and Oesterreicher (1985), its use for research cannot be the description of registers themselves, but rather the control of certain properties of texts during corpus compilation and statistical analysis in the context of the analysis of other linguistic phenomena. This is the idea of practical implementation through automatic analysis. The topic is taken up again in Section 2.3, using the current version of GiesKaNe with 24 texts and about 353,000 GiesKaNe tokens, 343,000 DTA tokens or about 297,000 words. The topic of GiesKaNe's tokenization is taken up again in Section 2.1.1.

The project must also take into account the object language level of historically variable language data as well as the metalinguistic discussion about suitable grammar models. With regard to the latter, current considerations on convergences and complementarities between projectionist and constructionist grammar models could be relevant (see, for example, Jacobs 2008; Welke 2011; Engelberg et al. (ed.) 2015). We choose a descriptive perspective without any claim to theory neutrality. GiesKaNe uses the pattern of alternating annotation of forms or constituents in nodes and syntactic functions in edges established for German-language treebanks (Krause & Zeldes, 2016: 129). Rehbein (2010: 231 f.) describes this development towards the hybrid approach. As Dipper and Kübler (2017: 600) note, NeGra is already based on the hybrid model and Dipper (2015: 547) points out that the two historical German treebanks available at the time, the 'Deutsch Diachrone Baumbank' (Hirschmann and Linde, 2010) and the Mercurius treebank (Demske, 2007), also use this scheme. A decision for one or the other perspective from the aspect of derivability or the question of what is more important, as briefly presented in Nivres (2008: 230), is not necessary. The grammatical model is Ágel's (2017) grammatical text analysis ('Grammatische Textanalyse', GTA). Central aspects are discussed in Section 2.1.3 in comparison to de facto standards for contemporary language treebanks. The GTA uses a syntax whose analysis is not directed from the word to the sentence and possibly to the text, but vice versa starts with the text. In the case of conceptually oral texts in New High German, such as correspondence from the Franco-Prussian War, this direction of analysis is also very practical if the segments of the text cannot be systematically accessed via its surface, but a solid parser analysis can be achieved through manual segmentation (see Section 2.1.2 and 2.1.3). In his critical discussion of Ágel's GTA, Welke (2020) emphasizes intersections with construction grammar and the orientation towards language use. Holler (2020) is critical of the GTA as grammar. However, she rightly concludes that this approach is not aimed at grammatically correct utterances that can potentially be produced by virtue of competence, but rather, in a performative manner, at real texts that have already been produced. This fits with the profile-forming aim to represent the entire range of concrete linguistic phenomena on the text surface in a grammar, which – according to Holler (2020) – makes the GTA appear advantageous, especially for corpus linguistic projects. In addition, this direction of analysis means that fewer assumptions are made as the proximity to the text surface increases so that "historical data speak for themselves" (Davidse and De Smet, 2021: 219). However, in GiesKaNe the level of PoS tags is seen primarily as a useful addition to the treebank when describing the contextual conditions of a phenomenon. This idea of 'additive' rather than 'consecutive' annotation is briefly discussed in Section 2.1.3. Emmrich and Hennig (2023) discuss the annotation concept in detail using the example of the annotation 'focusing modifiers' ('Fokusglied') rather than focus/degree particles ('Fokuspartikel') in GiesKaNe. Here, for example, its syntactic annotation in the treebank is supplemented at the level of the parts of speech by specifying a so-called donation PoS value. This is comparable to a strategy in the historical PoS tagset HiTS (Dipper et al., 2013), in which a distinction is made between the annotation of the lemma and that of the token. The broader topic of tagsets is addressed in Section 2.1.4. GiesKaNe needs additive annotation to complement the treebank. This contrasts with the comparability with other historical corpora that use the HiTS-PoS tagset. But GiesKaNe, as a complex treebank, must respond flexibly to the special requirements for the dynamic of change in New High German by fully utilizing the resources of corpus-linguistic annotation and keeping possibilities open for further developments and additions. The special requirements for the dynamic of change in New High German arise primarily from the socio-culturally induced transition from a horizontal to a vertical organization of the variety spectrum. Reichmann (1988) describes this process with the term 'Vertikalisierung des Varietätenspektrums' (Engl.: verticalization of the variety spectrum), i.e. from a social and spatial juxtaposition of varieties to a structure of varieties based on the model of a standard written language (cf. Emmrich & Hennig, 2022). The position between the Early New High German (reference corpus of the DDD family) and contemporary German, in addition to the extensive DTA for New High German, is accompanied by further challenges. Some of these challenges arise from the contrast between the need for standards and the demand for flexibility, requiring a balance that accommodates both consistency and adaptability in linguistic analysis and annotation. This question forms the overarching and connecting theme of Section 2.1.

---

[14] DWDS – Digital Dictionary of the German Language. The word information system for the German language in its historical and contemporary contexts, edited by the Berlin-Brandenburg Academy of Sciences and Humanities. https://www.dwds.de/
[15] As a guideline for the size of the text segments, we use the number of morpho-syntactic words instead of tokens, as the latter can vary depending on differences in punctuation.

Particularly in some conceptual oral texts, questions arise about the concept of sentences (2.1.2) and the corresponding segmentation and also about the relationship between automatic and manual work steps. Then it generally makes sense to normalize the text, also in preparation for the automatic analysis (2.2.1, 2.2.2). The treebank or rather the syntactic analyzes require a finer tokenization that differs from the orthographic word (2.1.1). In all of this, the surface of the text must remain 'intact' (a.o. Leech, 1993: 275). The broader topic of text preparation is therefore not addressed in the sense of a chronological structure of the workflow in order to form meaningful thematic blocks. This also applies to the automatic analysis, which shapes the project in all areas. Therefore, relevant topics are distributed across the subsections and not bundled together. The subtopics include (as far as possible) a study for application in GiesKaNe. In addition to the quality of the automatic analysis steps (e.g., parser and PoS tagger in 2.2.3.1) in a workflow that alternates between human actions or competencies and machines (2.2.2), the Inter Annotator Agreement (2.2.3.2) is also a central sub-topic. In this context I will briefly touch on the topic of the annotation tool used, which in our case is as simple Spreadsheet (MS Excel) (2.2.4), which – developed as an emergency or transitional solution – quickly proved itself and established itself.

Further considerations also include that the texts in GiesKaNe, as syntactically deeply annotated section of a text (about 12,000 morphosyntactic words), have to be related to complete texts in the DTA with its TEI-compliant preparation[16]. The selection of texts is based, if possible, on the texts of the DTA in order to support this connection; Especially when it comes to texts of immediacy ('Nähetexte'), new texts have to be incorporated.[17] The question of how well the selected text sections represent the entire texts whose metatag data they claim is dealt with in Section 2.3.

Basically, as explained, the article outlines the workflow in GiesKaNe, but also attempts to extensively integrate the relevant literature across topics and incorporate the current state of the art in automatic language processing and supplement it with empirical values from GiesKaNe. The roles of the GiesKaNe corpus mentioned at the beginning give rise to leading topics such as the relationship to other corpora and standards, that of research interests in the project and benefits for the entire community, as well as the relationship between humans and machines and everything, of course, against the background of effort and benefit.

## 2. GiesKaNe within different areas of tension: annotation scheme, text preparation, workflow, text compilation

As already mentioned, with the aim of creating a reference corpus, the community idea is at the center of the corpus compilation right from the start. And the same applies to annotations. They must also be seen in the tension between internal, more concrete research interests and external, more abstract use cases (cf. Emmrich & Hennig, 2022). In the recommendations on data technology standards and tools for the collection of language corpora from the German Research Foundation expert committee 'Linguistics' (DFG Fachkollegium Sprachwissenschaften, 2019)[18], it is recommended to use standards at least as a starting point, provided that they can be usefully applied for the respective research purpose. TIGER (Albert et al., 2003) is named as the 'de facto standard' for syntactic annotation, and the Stuttgart-Tübingen Tagset 'STTS' (Schiller et al., 1999) is named as the standard for morphosyntactic PoS tagging. Leech (2004: 30) describes de facto standards as "some kind of standardization that has already begun to take place, due to influential precedents or practical initiatives in the research community [...] in a bottom-up manner." He describes the sometimes conflicting interests mentioned here as follows:

> "De facto standards encapsulate what people have found to work in the past, which argues that they should be adopted by people undertaking a new research project, to support a growing consensus in the community. However, often a new project breaks new ground, for example with a different kind of data, a different language, a different purpose those of previous projects. It would clearly be a recipe for stagnation if we were to coerce new projects into the following exactly the practices of earlier ones. Nevertheless it makes sense for new projects to respect the outcomes of earlier projects, and only to depart from their practices where this can be justified." (Leech, 2004: 31)

A different kind of data, a different language, a different purpose: This can be concretized with the questions of whether, for example, conceptual oral everyday texts ('Nähetexte') can be adequately captured using TIGER (Albert et al., 2003) or whether TIGER, developed on the basis of contemporary language newspaper texts, can do justice to the phenomena of New High German or whether one can compare the purpose of the grammatical foundation of a reference corpus for New High German as in GiesKaNe with that of the TüBa-DZ (Telljohann et al., 2017), which is the second (major) constituency treebank of contemporary German and dispenses with crossing branches, "since TüBa-D/Z was intended to be used for parser training." (Meyers et al., 2006: 51) And as Meyers (2009: 106 f.) points out, consensus is difficult to quantify and "the descriptive-adequacy constraint provides important reasons to reject particular 'standard' analyzes when they are inadequate for describing particular data." With reference to some comparable historical German corpora, Odebrecht et al. (2017: 714) states accordingly: "Due to the different research questions, language periods and text genres these corpora use different annotation schemes, different normalization rules." Stefanowitsch (2020: 123), in turn, focuses on the need to keep the different research projects in a certain area comparable. He particularly emphasizes de facto standards in PoS tagging and parsing as easily applicable to new data. But he also has to admit that such annotation schemes, if they exist, often "will not be suitable for the specific data we plan to use, or they may be incompatible with our theoretical assumptions". Emmrich and Hennig (2022) also emphasize the fact that "all or nothing" applies when applying a standard. Of course, a de facto standard is not an unchangeable norm (a.o. Leech, 1993: 275), but interoperability (cf. Kübler et al., 2008) and usability are based on the fact that an expectation is met and that would make selective deviations all the more problematic. This could be counteracted with extensive documentation in the sense of 'caveat emptor' (Leech, 1993). But then at least the usability would be significantly reduced. Based on the scientific originality of the research project, the subject concerns a research gap and is at least partially unknown. Against this background, 'all or nothing' takes on even more weight. Interoperability as well as usability speak for the use of a standard. Dimensions of language variation in the sense of 'Nähe und Distanz' (Koch & Oesterreicher, 1985) and/or Coserius' distinction between diaphasic, diastratic, diatopic[19] conflict with the concept of standard in this all-or-nothing sense and require flexibility. In his handbook article, Nivres (2008: 226) speaks of a debate that is still ongoing: "It is still a matter of ongoing debate to what extent it is possible to cater for different needs without compromising the usefulness for each individual use, and different design choices can to some extent be seen to represent different standpoints in this debate." As we use Ágel's model of GTA (2017), a more recent but flexible grammatical model, in an attempt to meet the needs of new data, different language levels and other research interests, the obvious advantages of using standards must be taken into account in other ways. As in Meyers (2009) with reference to the study in Meyers et al. (2006), converting annotation can be a starting point for this problem. Emmrich and Hennig (2022) illustrate the possibility of a supplementary derivation of a standard based on existing annotations. This idea is addressed in Section 2.1.4 using the example of PoS tagging with HiTS, a variant of STTS for historical texts. GiesKaNe's role as a reference corpus, as a historical corpus and as a treebank was highlighted in the introduction and based on these 3 points, more concrete topics of our practical project work were addressed. These will be fleshed out in further sections.

In Section 2.1, the annotation scheme is introduced with reference to the grammar model of the GTA. The topic is characterized by references to other corpora and de facto standards in the dimension of interoperability. It is contrasted with the research interest in GiesKaNe and the granularity of the annotation scheme. The sub-topic of text preparation is also touched upon here through questions of tokenization, which is essential for the later annotations and must be included in the comparison of the annotation schemes. This is particularly true for historical corpora, and this also applies to the recognition of sentence boundaries and a corresponding segmentation of the text. Both 'preparation steps' are the basis for further comparison of GiesKaNe's annotation schema and possible de facto standards. The problem of a 'new' research interest or scientific progress on the one hand and the use of de facto standards with the advantage of interoperability and greater usability on the other hand is taken up again in a practical study using the derivation of HiTS. HiTS is a de facto standard for PoS tagging of historical texts in German. This is derived from existing manually corrected annotations in GiesKaNe (cf. Emmrich & Hennig, 2022). With reference to Emmrich and Hennig (2023) I will briefly discuss the idea of 'additive' rather than 'consecutive' annotation in this Section, too.

---

[19] See the English overview in Kabatek (2023); on the relationship between all these dimensions: cf. Fischer (2008).

Section 2.2 then discusses the workflow in which human and automated analyzes are combined. In addition to determining the macro units of the text (especially the sentence boundaries), a sub-topic of text preparation is again fundamental here: the normalization of the texts as the basis for machine analysis with a contemporary language model. A dependency parser is central to the subsequent automatic analyses. Its training data is obtained from a conversion of the constituent structure, and its results are then converted back into a constituent structure so that it can be manually corrected again in a spreadsheet. In addition to questions about parser evaluation, the IAA is also discussed here. How complex syntactic annotations are made in a spreadsheet is also addressed to demonstrate that a project like GiesKaNe does not have to produce a special tool that requires maintenance. However, I will also discuss some ideas about the one annotation tool that still needs to be designed.

Section 2.3 discusses a method for implementing the model of 'Nähe und Distanz' (Koch & Oesterreicher, 1985) in automatic analysis (cf. Emmrich, 2024). For the GiesKaNe corpus, a rough binary categorization in the meta data can be replaced by a scalar ranking. However, the main benefit may be different: When selecting texts, automatic analysis is crucial because there are no human resources available for this, apart from looking at the communication conditions, social circumstances and topics. It is therefore a question of how the text selection by experts can be supplemented by an automatic analysis that provides the internal linguistic features of the text and can relate them to the external communication conditions. In addition, this section also tests how well the text segments contained in GiesKaNe represent the entire texts contained in the DTA. For a treebank, a corpus that has to limit the text length due to in-depth analysis, this question is even more important. Ultimately, some partial analyses in this article will also demonstrate that a machine-based classification of texts on a scale of conceptual orality and literacy, carried out prior to the text entering the workflow, could offer valuable insights into the labor intensity of a text and the time required for the project's workflow processing.

## 2.1 The annotation scheme: From tokenization to sentence boundaries, its relation to the grammatical model and to other corpora or standards and an attempt to meet multiple interests

This section was introduced at the beginning with reference to the advantages and disadvantages of using a de facto standard. In this respect, the grammar model of Ágel's GTA used in GiesKaNe is compared with the annotation schemes of comparable corpora. The large German-language treebanks for contemporary German TIGER and TüBa-DZ are particularly suitable for a comparison. Not fundamentally, but selectively all the more interesting are comparisons to the Universal Dependencies – especially from the perspective of extensions for historical German-language texts proposed by Dipper et al. (2024) – as well as the Mercurius treebank (Demske, 2007), Fürstinnenkorrespondenz (Engl.: Princess correspondence) (Prutscher & Seidel, 2012), the projects in 'Deutsch Diachrone Baumbanken' (Hirschmann & Linde, 2010), Corpus of Historical Low German (Booth et al., 2020) and ReF.UP[20] (Wegera et al., 2021; cf. Ortmann, 2021). Of the corpora mentioned, the Indiana Parsed Corpus of (Historical) High German (Sapp et al., 2024) stands out as a parsed Penn-style treebank.

The references to the more flatly annotated historical corpora are also relevant: With reference to the DDD[21] family (Lüdeling et al., 2005), GiesKaNe closes the gap between Early New High German and contemporary German within the framework of the manually morpho-syntactically annotated corpora. For this period of New High German, the extensive DTA is also available as an automatically PoS tagged corpus. Compared to the DTA, GiesKaNe fits into the corpus infrastructure as a treebank, especially with deeper manual annotation. The DTA as well as the large contemporary language corpora – such as the DWDS and DeReKo – which represent the other point of connection for GiesKaNe, rely on automatic PoS tagging according to the STTS (Schiller et al., 1999). A comparison is particularly possible using the HiTS (Dipper et al., 2013) tagset for historical language, which is derived from the STTS.

Since they form the smallest and largest analysis units of the treebank at two end points and thus represent the basic units of analysis, I will start with the discussion of tokenization and sentence terms.

## 2.1.1 Tokenization: The basis of grammatical analysis as part of the annotation scheme

Considering tokenization as part of the annotation scheme is less common or not necessary for contemporary language corpora, but, as Dipper et al. (2013: 87) point out in the context of the presentation of HiTS, it makes sense in the case of historical corpora: Neither spaces indicate word boundaries nor punctuation marks indicate sentence boundaries, which Demske (2007: 94 f.) points out in her comments on the Early Modern German Mercurius treebank, too. Kroymann et al. (2011: 5) address the difference between graphic and lexical words and make it clear that tagsets that were developed for modern language levels are not easily transferable to older language levels. As Odebrecht et al. (2017) note with reference to their Ridges corpus, segmentation should be seen as an interpretation of the primary data, and this can vary – depending on the research question and the assignment criteria (cf. Lüdeling, 2011). As Denison (2013: 17; cf. Durrell, 2015) notes, you get in a corpus (only) what the theory can handle. In this sense, tokenization must be seen as part of the annotation scheme and therefore this processing step must also be included in considerations regarding the application of a de facto standard. Because if you look at tokenization in TIGER (Albert et al., 2003) and TüBa-DZ (Telljohann et al., 2017), this is virtually a prerequisite. The step is not addressed in the annotation scheme for TIGER as well as that for TüBa-DZ. In their comparison between TIGER and TüBa-DZ Dipper and Kübler (2017) say about TIGER: "As the very first step, the texts of the corpus were tokenized. The tokenized sentences were proofread once by the annotators". This also applies to the tokenization in TüBa-DZ. Dipper (2013) points out in her comments on HiTS that clitized forms are often used also analyzed as independent tokens, for example in the TIGER corpus: *gibt's* (Engl.: there's). Of course, a different tokenization does not exclude the use of an annotation scheme, but comparability as a specification of interoperability and usability is reduced here for the first time.

This topic is dealt with throughout the historical corpora or with reference to them. Odebrecht et al. (2017: 699) choose the approach of multiple tokenization: "Our goal is not to enforce a single minimal tokenization to which the other tokenization refer, but to allow conflicting segmentations". Such an approach is discussed in Sauer and Lüdeling (2016), including the possibility of being able to react flexibly to new requirements. Another solution is put forward by Krause and Zeldes (2016: 132) to address the tension between the correct representation of the physical manuscript and the text with its focus on substantive aspects. They propose sub tokenization as an alternative to the 'word' level. Durrell (2015: 26) shows that the concept of the word is fundamental for tokenization, and he illustrates this using the example of 'on behalf of', which is always tagged in two ways in the BNC – not just as a complex preposition. With reference to Manning and Schütze (1999), Chiarcos et al. (2012: 55) also point out that the concept of the token is neither trivial nor the same as the word and advise against a uniform definition: "Instead, we argue that the definition of the term 'word' depends on the research questions or applications of interest." Zeldes (2021: 53 f.) summarizes the relationship between word and token as follows:

> "Although a working definition of 'tokens' often equates them with 'words, numbers, punctuation marks, parentheses, quotation marks, and similar entities' (Schmid, 2008: 527), a more precise definition of tokens is simply 'the smallest unit of a corpus' (Krause et al., 2012: 2), where units can also be smaller than a word, e.g. in a corpus treating each syllable as a token. In other words, tokens are minimal, indivisible or 'atomic' units, and any unit to which we want to apply annotations cannot be smaller than a token."

Since at least the text surface must be preserved despite a more complex tokenization, HiTS or the DDD corpora, which – at least partially[22] – use this tagset[23] use two levels of text representation: a diplomatic one, i.e., close to handwriting layer and a layer with corresponding tokenization. In GiesKaNe this is guaranteed via the DTA tokenization layer ('DTA_token') in addition to the tokenization for syntactic analysis (token). GiesKaNe's guidelines for text preparation[24] state: The tokens are split if a syntactic function at sentence or phrase level is to be assigned to a GKN token created in this way. Even if this may initially sound like individual decisions, there are – as the compilation by Durrell (2015: 25 ff.) shows – case groups. Schröder et al. (2017: 43) only see the possibility of manual tokenization in the reference corpus Middle Low German/Lower Rhine (REN) from the DDD family. In GiesKaNe automatic rule-based separations are made in the areas of prepositions (contraction of preposition and definite article or prepositional adverbs as combinations of *da(r)* or *wo(r)* or *hie(r)* +

---

[22] Because the usage here is not entirely uniform (cf. Emmrich & Hennig, 2022: 211 f.).

[23] Dipper & Kwekkeboom (2018: 97) note that the texts in ReM (unlike in ReA) were diplomatically transcribed, i.e., editions were only used to support this process. In ReF too, all texts were diplomatically transcribed (up to 20,000 word forms) (cf. Dipper & Kwekkeboom, 2018: 99). Diplomatic transcription is primarily discussed in the context of REM.: Dipper, 2015: 526 f., 2021: 148 f., Klein & Dipper, 2016, Petran et al., 2016, Dipper & Schultz-Balluff. Dipper (2015) explains this in detail in the presentation of HiTS: Basically, it is a transcription that is as close to handwriting as possible. A detailed discussion of tokenization also takes place as part of the basic presentation of the plan for DDD in Lüdeling et al. (2005). For ReN Schröder et al. (2017), Peters and Nagel (2014) as well as Peters (2017) mention a diplomatic transcription, but they do not define this practice in more detail. The concept is also addressed in Hirschmann and Linde (2010) in their annotation guidelines for the German Diachronic Treebank (DDB).

[24] https://gieskane.com/wp-content/uploads/2023/07/richtlinien-fuer-die-textvorbereitung.pdf

preposition), pronominal cliticization (*ichs*, *sichs*, *hastu*; see end of section), particle verbs like (*fort-bestehen*, *los-laufen*, *weg-tragen)* and orthographic words with 'tokens' like *einige*, *mehr* or *mal*, which is documented in the tokenization guidelines. Here, Stefanowitsch (2020: 85), who like others emphasizes that tokenization is also the result of decisions, can point to the relevance of an operational definition: "Put simply, an operational definition of a construct is an explicit and unambiguous description of a set of operations that are performed to identify and measure that construct." (Stefanowitsch, 2020: 77) But this is about more than Leech's (1993) 'caveat emptor'. The term sub tokenization by Krause and Zeldes (2016) is, in my opinion, not optimally chosen – taking into account the definition as an 'atomic unit' according to Krause et al. (2012) or Zeldes (2021) –, but it does indicate that the word-term has developed great relevance, also from the aspect of usability: here, one should not equate the technical possibilities of a multi-level annotation with its actual use. A query[25] in ANNIS for prepositional objects in the form of a prepositional adverb *darauf* for example makes it clear that the compatibility of the orthographic word term and tokenization for syntactic analyzes leads to the token becoming a kind of mediating unit between the treebank (in a narrow sense) and the span annotations as in the DTA tokens. This must be taken into account in the query in that two nodes linked by a prepositional-object edge must be searched for, but the child node is not the word form *darauf*, but rather includes two tokens that have 'identical coverage' ( _=_ ). This is counterintuitive, or at least counterintuitive to many potential users. It has to be documented and ultimately taken into account and such a simple search increases the level of complexity of the query to a moderate level. But especially against the background of preserving the text surface against equally necessary but not precisely foreseeable grammatical analysis units, this additional effort seems justified, but this sub tokenization should be carried out in a systematically clear manner.

At this point, I will focus briefly on the automatic implementation of tokenization in GiesKaNe. While closed classes (contractions of preposition and article as well as prepositional adverbs) and some patterns of cliticization can be split by regular expressions, the tokenization of the particle verbs (in a broad sense) is based on an analysis of the DWDS corpus, in which 21,670 particle verb types and their frequencies were recorded via the orthographic property of the spelling in infinitive constructions (*loszulaufen*), where the combination of a particle and a simple verb form is 'split' by *zu*. In this way, the actually occurring combinations of stem verbs and separable particles could be identified efficiently. Using these findings, 3,834 tokens (2.57%) had to be split in a study from an earlier project phase with manually revised texts (around 150,000 tokens, 10 text: 4 x 17th, 3 x 18th and 3 x 19th century). In the automatic implementation, the recall is 0.81 and the precision is 0.97. Since the product of the automatic text preparation is a spreadsheet with the tokens in column A, to which DTA tokens and normalization a.o. refer as connected cells in other columns (Fig. 2).

Fig. 2: Automatic preparation of a GiesKaNe text for further manual processing in MS Excel based on the DTA version

There are various reasons for this recall. What is crucial is the fact that automatic tokenization is done in one step with automatic normalization, so that both can be corrected manually in one step. In this regard, tokenization depends heavily on the quality of normalization and of course cannot yet build on automatic PoS tagging: About 63% of the false negatives come from 3 texts from the 17th century. Here the chars and orthographic variants are less predictable. An adaptation to the numerous variants would be disproportionate due to the lack of systematicity or the possibility of applicability to 'comparable' texts. The number of false positives would then also increase. The starting point for this automatic text preparation was the idea that as few new lines as possible had to be inserted manually and that the existing annotations adopted from the DTA should not need to be merged manually (Fig. 2, 'DTA_Orth' = merged cells for the orthographic sentence or 'DTA_Token' = merged cell for orthographic words). A high number of false positives was deliberately counteracted because splitting the cells manually would have been just as time-consuming. One central finding from this study that led to an improvement of the algorithms precision is that in 180 prepositional adverbs an *r* is realized even though the preposition that followed *da* (there) and *wo* (where) begins with a consonant.[26] This, in turn, is not found in current standard German, but is discussed against the background of language change processes by Negele (2012) with reference to Fleischer (2002) for certain conceptually oral texts/texts of immediacy (Koch & Oesterreicher, 1985). But the area of pronominal cliticization can also be structured even better in the future after analysis of types and token. While an -s at the end of the token offers an obvious starting point, there is a lot of variation in the relationship between types (114) and tokens (200) and only a quarter (46) of 167 orthographic words ending in -s do end in -'s. Only 12 of these tokens to be separated appear more than twice, then accounting for 75 cases: *ichs* (ich es, Engl. literal: I it), *sichs* (sich es, Engl. literal: itself it), *ifts* (ist es, Engl. literal: is it), *gings* (ging es, Engl. literal: went it), *hastu* (hast du, Engl. literal: have you), *fichs* (sich es, Engl. literal: itself it), *gefchichts* (geschieht es, Engl. literal: happens it), *ich's* (ich es, Engl. literal: I it) , haftu (hast du, Engl. literal: have you), *mir's* (mir es, Engl. literal: me it), *machts* (macht es, Engl. literal: make(s) it), *werdens* (werden es, Engl. literal: will it). This leaves a lot of room for variation – as you can see, in the orthography as well as in verbs, pronouns and other bases: *soldens*[27] (normalization = sollten es, Engl.: should it), *Richteuch* (Richt euch, Engl.: Align yourselves), *Sieheftu* (Siehst du [nicht täglich]…, Engl.: Don't you see every day…), *behaltu* ([das andere] behalt du, Engl.: [the other,] you keep), *folgeftu* ([warum] folgst du, Engl.: [why] do you follow), *fraßens* ([sie] fraßen es, Engl.: [they] devoured it ), *gaber* (gab er, Engl.: he gave), *hätt's* (hätte es [nicht in meinem Kopf gelegen, …], Engl.: had it [not been in my mind, …]), *ifter* (ist er, Engl.: he is), *klingeter* (klingt der [Kanzler], Engl.: the chancellor sounds), *lernetens* (lernten es, Engl.: learned it), *liffens* (liefen es, Engl.: ran it). This perhaps illustrates why one should compromise on tokenization or recall while simultaneously normalizing.

## 2.1.2 Sentences and other (macro) units of text analysis

As mentioned, in the annotation scheme TIGER and TüBa-DZ expect not only tokenization but also segmentation of text into sentences. A comparison of the two annotation schemes from the perspective of application in parsing by Rehbein and van Genabith (2007: 631) shows that with 17.6 or 17.27 words per sentence, there should be hardly any differences in this concept, which should correspond to that of the orthographic sentence: "They [(newspaper articles) are preprocessed into syntactic units delimited by punctuation marks (. ? ! ; - ... /) for which specific rules demand or forbid segmentation." (Telljohann et al., 2017: 22) While SIMPX (simplex clause) represents the central analysis unit of TüBa-DZ, it is given the value VROOT together with the punctuation marks. So the definition is not entirely correct, but that should not really be a problem. However, a note by Petran (2012: 78) highlights a problem: "Due to the recursive nature of clauses, their extraction from TüBa-DZ is not as straightforward." He refers to the fact that one node SIMPX captures units that represent both paratactically linked/coordinated units and the result of the link in a single context of an orthographic sentence. What initially seems unfortunate from a usability perspective must ultimately be considered in several dimensions. Later I will discuss the opposite case and critical aspects of the GiesKaNe tagset, in which different sentence types can be precisely differentiated. However, further optimization can be done here, too.[28]

Here, what, for example in TüBa-D/Z, is problematic from the perspective of querying in terms of usability also raises the central question about the concept of sentence and the real topic of this section. Because what is not discussed in contemporary language annotation schemes is – as numerous articles will show – central to historical grammar and historical corpus linguistics. In the context of the Mercurius treebank for Early New High German, Demske (2007) points out with reference to Stolt (1990) that sentence boundaries are only very unreliably indicated by punctuation marks. Biber and Conrad (2009: 154) note, for example, that in 18th century novels "the full-stop (.) functions almost like a paragraph marker rather than a sentence marker". Eckhoff and Berdičevskis (2016: 66) notice an opposite tendency for Old Slavic texts: "Old East Slavic texts generally use syntactically motivated punctuation, but the punctuation usually indicates smaller syntactic units than the sentence." Booth et al. (2020: 773) also have problems determining main and subordinate clauses in their Penn-style Treebank of Middle Low German because "sentential punctuation and capitalization are often absent, or used in a way which does not systematically distinguish main and subordinate clauses." Walkden (2016: 565) remark in relation to Sievers' (1878) edition of the epic poetry Heliand in the Helipad corpus of Old Saxon that "only one punctus" is used completely arbitrarily cannot be surprising given the genre. Petran (2012: 75) generalizes: "Syntactically motivated punctuation was not used until well in Early Modern times for European languages [...]." In addition to a study on the use of punctuation in historical texts Odebrecht et al. (2017: 715) confirm for the German language: "In former stages of German, there was no binding orthographic norm for punctuation in the written language (Hochli, 1981; Simmler, 2003; Nerius, 2007)." Their study shows that slashes were widespread before 1700 – also at the same time as commas – and then disappeared after 1700 in favor of the comma.[29] The practice of automatic sentence boundary recognition based on punctuation in New High German texts by Sapp et al. (2024: 9225), who present the Indiana Parsed Corpus of (Historical) High German, invites the question of what consequences follow from the fact that they themselves note that their approach "can be problematic for early texts". The problem is ultimately summarized by Davidse and De Smet (2021: 219):

> „Finally, any effort at creating richly annotated corpora runs the risk of obscuring existing patterns in the data. It is not obvious, for instance, that Old English authors had a concept of sentences that is exactly comparable to the notion of sentence assumed by the formal theory underlying the parsing in YCOE – Old English punctuation, in any case, suggests otherwise (Fischer et al. 2017:163). In other words, in the end it is again the researcher who should always be wary of any prior assumptions and who should try to find ways to let historical data speak for themselves. At the same time, the ideal practice for corpus compilers is to strive to give end users access to original spelling, punctuation and even text layout. As is the case now for many online text archives, users can only benefit from being able to consult high-quality images of the original manuscripts or prints in the corpus.“

---

[26] https://annis.germanistik.uni-giessen.de/#_q=RFRBX3Rvaz0vKGRhfHdvKXJbXmFlaW91XS4rLyAmIHdhPSJwciIKJiAjMSBfaV8gIzI&ql=aql&_c=R2llc0thTmVfdjAuMw&cl=5&cr=5&s=0&l=10.

[27] DTA_tok="soldens", https://annis.germanistik.uni-giessen.de/#_q=RFRBX3Rvaz0ic29sZGVucyIK&ql=aql&_c=R2llc0thTmVfdjAuMw&cl=5&cr=5&s=0&l=10

[28] I touch on the distinction between macro, meso and micro levels, which is central to Ágel's GTA (2017), on the one hand, and the concept of recycling on the other.

In this sense, it does not seem promising to apply contemporary grammar models to historical language levels without the possibility of flexible adaptation or to make orthographic sentences the basis of analysis if they were unable to be fundamental for communication at this time and if they cannot be systematically determined today. And in this context, historical texts can be preliminarily analyzed using contemporary language parser models (see Section 2.2.3.1), but they absolutely require manual adjustment. The relationship between GiesKaNe and the DTA, whose scans of the texts are also occasionally used in practical analysis work in GiesKaNe, should also be emphasized here.

But the problem in this area of tension between contemporary language and historical language levels and de facto standards or flexibility must also be understood in another way. It is not just a matter of fact that sentence boundaries in historical texts cannot systematically be determined by machine. The focus must be on the concept of the sentence. Pietrandrea et al. (2014)[30] address 3 starting points for sentence definition, which in my opinion can be summarized as follows: the speaker's unit of communication, an orthographic or prosodic unit and a syntactic sentence concept. With reference to Nivre (2008), who questions whether "to what extent the annotation schemes developed for written language are adequate for the annotation of spoken language"[31], Pietrandrea et al. (2014) provide a bottom-up annotation without prior analysis of the units. They particularly point out that for semi-automatic analysis, the parser requires pre-segmentation. In my opinion, the crucial question is which sentence concept is the starting point. Pietrandrea et al. (2014) define 'clause' with Berrendonner (2002) as "the projection of a syntactic dependency tree whose head does not dependent on any other word in the sequence".

Ágel's grammatical textanylsis (2017), which is used in GiesKaNe, is a descendant, top-down syntax of German applied to historical texts. But the apparent contradiction between the criticism of an orthographic sentence concept and the analysis process in GiesKaNe does not exist, because Ágel's GTA (2017) is not based on the orthographic sentence concept and its orientation on punctuation, which – as mentioned – is hardly usable for historical texts anyway: In contrast to the syntactically indefinable orthographic sentence, the grammatical sentence represents one of the three macro-units of the text (along with units of cohesion and non-sentences). A grammatical sentence contains a single main predicate that creates a scenario via valence. This does not contradict formal realizations, such as coordination ellipses. Predicate and complements realize the the content of the sentence and these sentences have a relative autonomous coding. As a textual unit with its inherent autonomy, the concept of the grammatical sentence differs from the broader notion of a clause, which can also appear embedded as a subordinate clause.

At this point I return to Petran's criticism (2012: 78) of the recursivity of the sentence structures in TüBa-D/Z. Because it can be mentioned here that the tagset of GiesKaNe v0.3 contains 13 types of clauses (e.g. *ns*, *ns_un* and *ik* for subordinate clause, independent subordinate clause and infinitive construction), which make it clear that it is not a macro unit of the text (*s*), but an embedded sentence with a syntactic function in the macro sentence unit. The tags also encode further information about its formal properties that do not need to be reconstructed. However, the central concept, 'from predicate follows sentence', is not made transparent here. The recursivity of language is addressed in the GTA (Ágel, 2017) through the concept of recycling. Applied here, it also means that 'if you know the grammatical description of the sentence, for example, you don't have to reinvent that of the subordinate clause' (Ágel, 2017: 718). The sentence represents a text-grammatical value, while subordinate clauses represent possible forms (arguments) of meso- and micro-functions (Ágel, 2017: 291). The inherently more complex topic of the three levels of grammatical description in Ágels GTA and the concept of recycling is addressed again in Section 2.1.3 in connection with the granularity of the annotation scheme.

---

[29] It is interesting that Vorbeck-Heyn (2010: 823 f.) states for a Bible text by Luther (published in 1583) that the beginnings of sentences are marked using virgel + majuscule, subordinate clauses within the overall sentences and other parts of sentences are marked by virgel + minuscule. The distinction between constituents/functions within or of clauses ('Satzglieder') on the one hand, and those within phrases ('Gliedteile') on the other hand, is discussed in Section 2.1.3.
[30] Pietrandrea et al. (2014) discuss the use of different units of text structuring. Hirschman et al. (2007) also propose the combination of multiple annotation levels.
[31] Dash (2021: 4 f.) uses the term 'divisibility' and also emphasizes the fact that the concept sentence is difficult to apply to spoken language data, referring to the work of Wallis and Aarts (2006).

The corpus linguistic solution/consequences can be hierarchical or modular tags, as used primarily in PoS tagging.[32] But, as already stated by Skut et al. (1997:90) on the NeGra treebank: "However, there is a trade-off between the granularity of information encoded in the labels and the speed and accuracy of annotation."[33] This approach is discussed by Merten et al. (2021: 446), for example. They include the rough set theory (Pawlak, 1982) in their considerations and summarize that "due to limited capabilities in perception or measurement, certain elements […] might not be distinguishable".[34] Their study on annotation uncertainty in the context of grammatical change is taken up again in Section 2.2. The topic of sentence types in GiesKaNe is more about optimally preparing and using existing analyzes. This can be improved in version v.04. The following is about the macro unit 'sentence' i.e., sentences as units of the text.

If the units of the text are determined manually before a more complex syntactic analysis, this is neither contradictory nor a great deal of additional effort, because in communication we are trained to identify comparable units of syntax and semantics and other dimensions (somehow), which are then assigned a first analysis by an automatic parser. The advantage is that when segmenting, a person does not have to laboriously concretize this analysis. The syntactic parser, in turn, can work better with the grammatical sentence unit because it is based on a system that orthographic sentences sometimes lack. The parser shows its strengths when it comes to the mass of the individual hierarchical constituents. Humans, in turn, can then resolve ambiguous structures with reference to semantic and pragmatic features when controlling the parser analysis. (see Fig. 2.2.2, Section 2.2.3) Eckhoff and Berdičevskis (2016: 66) are guided by similar considerations:

> „Since we do not wish to use texts with editorial punctuation, and in many cases do not even have access to such punctuation, the final sentence division must necessarily be manual. However, if the sentence division is done manually before syntactic preprocessing, we would be doing double work and would be likely to lose any speed gain. Our syntactic preprocessing should therefore be performed before manual sentence boundary adjustment. This presents another problem: if the annotators change boundaries of pre-parsed sentences, the effort required to save the existing trees will cost more than annotating sentences from scratch, as demonstrated in pilot experiments."

But they seem to be limited by the annotation and corpus software used and its workflow that they ultimately perform parser analysis before manual segmentation. For their experiments on the question of facilitating manual annotation through prior parsing, they then only select text passages in which no sentence boundaries have to be changed manually. They then confirm the effect of the previous parsing in the sense of a significantly faster working time, but the question remains of course as to how much these text sections promote a correct parser analysis (for the role of immediacy and distance ('Nähe und Distanz', Koch & Oesterreicher, 1985) see Section 2.2.3).

I will close this section with a short excursus on a practical analysis, too. And I will briefly discuss the concept of the grammatical sentence and its analysis. We do not use the possibilities of automatic sentence boundary detection in the GiesKaNe workflow described. However, the parser analysis is based on manual segmentation. Like Farasyn et al. (2018: 286 f.) emphasize, this is highly useful for parser analysis and, in our opinion, a good investment of working time. However, the automatic determination of grammatical sentences is relevant for the automatic determination of a text's value of conceptual orality and literacy (Emmrich, 2024, see Section 2.2.3). Two models were trained with spacy's SentenceRecognizer based on the annotations in GiesKaNe_v0.3 (24 texts, each with 80% training, 20% testing): one for orthographic sentences (F-score = 0.94) and one for grammatical sentences (F-score = 0.45). For the latter, all punctuation marks were removed (such as in Petran, 2012) from the GiesKaNe token layer. The use is sequential and, for seven new texts that are only used for evaluation, it provides an average agreement of 77% in the assignment of the tokens to the sentences with the texts annotated in GiesKaNe. The mean values are between 0.86 (sd = 0.21) for Gellert[35] (fiction/prose, 1747) and 0.60 (sd = 0.28) for Bräuner[36] (science/medicine, 1714) and the proportion of identically segmented grammatical sentences is between 0.59 (Gellert) and 0.19 (Bräuner) (mean = 0.38). In principle, automatic segmentation into grammatical sentences as a basis for parser analysis seems conceivable. Two points are critical: On the one hand, in practice, unlike what is implemented here, grammatical sentences are not always smaller than orthographic sentences. The tokens of one grammatical sentence can therefore lie in two orthographic sentences. On the other hand, the texts sometimes differ significantly and even contrary to expectations: there are only 30 years between Gellert's novel with the best values and Bräuner's scientific writing with the worst. As a rule, however, the orthographic sentences (measured in tokens) are the more extensive units and I will illustrate this at the end of this section by comparing the orthographic sentence and the text segmentation units used in GiesKaNe, whereby the different texts in GiesKaNe are also taken into account. The challenge of defining grammatical sentences is not directly apparent from the given definition. Fundamentally, the concept requires a delineation of units of cohesion, non-sentences with which they form the units of analysis of the text, and ultimately orthographic sentences. Units of cohesion stand in the so-called intermediate space ('Zwischenstelle') between two sentences or non-sentences (Ágel, 2017).[37] As you can see from the query, these are primarily conjunctions, which link autonomous units of the text, not units within a syntactic structure of a sentence. Non-sentences, in turn, are a heterogeneous group.[38] They have no main predicate and no sentence bracket. This includes headings as well as exclamations, but also other phenomena like elliptical structures (cf. Hennig, 2011; Behr & Quintin, 1996) – as example 1 that is highly functional according to a potential predictive pattern X = Y.[39]

---

[32] Dash (2021: 28 f.) argues for a multi-level approach when it comes to annotations, and we also use a hierarchical tagset, whose atomic values can be queried independently. In PoS tagging, this at least reduces the reliance on regular expressions in queries. The system remains more open to combinations that were not originally present or intended in the annotation. As a de facto standard, the STTS (Schiller et al., 1999: 4) demonstrates less the modular nature of hybrid tags and more their rigidity. In contrast, approaches like HiTS (Dipper et al., 2013: 85) illustrate how these structures can be re-mobilized effectively.

[33] This also applies to automatic analyses, such as Zinsmeister et al. (2008: 766) with reference to Kilgarriff and Rosenzweig (2000), Bayerl et al. (2003), Kübler (2005) and Kübler et al. (2006). In this context, van den Bosch (2008: 859) also mentions taking into account the complexity of the source language, while Weisser (2022: 91) highlights the connection to the potential use of the corpus. Similarly, Zufferey (2020: 172) addresses these considerations.

[34] Flickinger et al. (2017) emphasize here the connection between human and automatic processing steps.

> **Example 1 (Anthus)[40]:**
> […] und so blieb nichts übrig, als durch Schaden klug zu werden; <u>Schade nur, daß Letzteres nicht immer auf Ersteres folgte.</u>
> <u>Engl.:</u>
> […] and so there was nothing left but to become wise through harm; <u>just a shame that the latter didn't always follow the former.</u>

Since no verb can create a scenario here, only forms and functions at the phrase level (below the verb level) are annotated. This also corresponds to the practice in TüBa-DZ (Telljohann et al., 2017: 29). A further analysis would require the distinction of certain types of ellipses (cf. Hennig, 2011; Behr & Quintin, 1996) and that would be even more difficult to implement with historical texts. The problem with this step becomes clear when compared to the TIGER scheme, whose analysis practice would no longer be transparent here: Although the problems of ambiguity are pointed out there, the rule is ultimately formulated that verbless sentences (e.g. in headings) should be supplemented in a meaningful way and then annotated in the normal way (Albert et al., 2003: 72).

If you look at the GiesKaNe corpus to see which patterns from these three (macro-)analysis categories (grammatical sentence, units of cohesion, non-sentences) only occur within an orthographic sentence, you get 616 types of patterns if you take the fixed position into account, 331 if it's just about the number and combination of units without positioning. Of 9,574 orthographic sentences analyzed[41], 4,202 were identical to the grammatical sentence. Less problematic for the parser are 1080 matches based on the sentence-cohesive unit/conjunction-sentence-pattern (e.g., s *and* s). In terms of grammar theory, the effects are problematic, but they are still systematic and controllable. 340 single cohesion units appear to be unproblematic, too. 853 combinations of two grammatical sentences without punctuation marks and conjunctions have to be recognized during segmentation. The statistical model that was trained to recognize the boundaries of grammatical sentences shows, with an f-score of just 0.45, that extensive manual correction interventions are required. This leaves around 32 percent for the more interesting matches with 234 hits for the most frequent and not yet that complicated pattern s + s + *and*[42] + s. Then there are s + s + s (213), non-sentences (186), s *and* s *and* s (169), s *and* s + s (118), *and* s + s (89), *and* s *and* s (88), s + s + s *and* s (64), s + non-sentence (47), s *and* s + s *and* s (40) and almost another 20 % for more complicated and much rarer patterns like those with discontinuous sentences. Example 2 shows a simple variant of this: There are 2 grammatical sentences at the beginning. Due to the pre-field of the second sentence, which is already marked with *es*[43], the word *vielmehr* (*maybe*) stands in between the grammatical sentences as a cohesive unit. A second cohesion unit *aber* (*but*) then stands 'before' a non-sentence (without a sentence bracket) *nicht allein dieß* (*not just this*), which is why a classification as 'preceding' can only be assumed as very likely, but not certain. At the end of the orthographic sentence there is a coordination of three grammatical sentences, in which greater cohesion arises from the fact that the second and third conjunctions begin with the finite verb and (elliptical) refer to the subject *er* (*he*).

> **Example 2 (Koralek)[44]:**
> [Über dies ist der große Otto aus seinem Posten aus getreten]ₛ , [vielmehr]_KG [es wurde ihm wegen seiner schlechten Schrift gekündigt]ₛ , [aber]_KG [nicht allein dieß]_NI , [er wurde krank]ₛ , [u.]_KG [mußte vielleicht 8 Tage zu Bette sein]ₛ , [u]_KG [muß sich jetzt noch erholen]ₛ .
> Engl.:
> [Furthermore, the great Otto has resigned from his post]ₛ , [rather,]_KG[45] [he was dismissed because of his poor writing]ₛ , [but]_KG [not just this]_NI , [he got sick]ₛ , [and]_KG [had to be in bed for maybe 8 days]ₛ , [and]_KG [now has to recover]ₛ .

On the one hand, this makes it clear that a syntactic analysis presupposes such segments as units. On the other hand, Example 2 shows the differences in the boundaries and variants of these units of the text. While problems are inevitably encountered in automatic analysis, these constructions could also be captured in manual annotation if the annotation scheme for child nodes of the orthographic sentence is differentiated. But here at least the question arises as to whether the initial orientation towards the orthographic sentence would still be justified. In contrast, there would be cases in which a macro unit of the text, as a syntactic-semantic subsystem of the text, would be divided into several orthographic sentences and could then no longer be analyzed at all. Such constellations as in Example 3 are rare, with 213 matches compared to 9,574 orthographic sentences, but they have a profound effect on the possibility of analysis.

> **Example 3 (Gessner)[46]:**
> Man wird es mir vergeben, daß ich mich deswegen nicht weitläuftig eingelaſſen habe. Weil es ohnehin bekannt genug zu ſeyn ſcheinet , daß man dergleichen Unternehmen von Scythen nicht einmal vermuthen , geſchweige denn glauben , kŏnne .
> Engl.:
> I will be forgiven for not getting involved in much detail because of this. Because it seems to be known enough anyway that such ventures by Scythians cannot even be suspected, let alone believed.

Especially in comparison to the English translation, the position of the finite verb *ſcheinet* at the end of the clause should be emphasized here, which, supplemented by the causal subjunctor *weil* (opposite, at the beginning of the clause), formally indicates a subordinate clause. An independent analysis of this unit therefore does not seem to be productive in New High German. But from the comparison of syntactic structure and orthography, further analysis possibilities arise in GiesKaNe. In principle, we can use the different annotation levels to well depict the problematic relationships between punctuation and orthographic sentences, on the one hand, and between this and the grammatical sentence, which have been discussed in detail in the literature.

---

[42] Here meant as a representative for all cohesive units.

[43] The placeholder *es* (similar to it) cannot be translated into English.

[44] GiesKaNe: https://annis.germanistik.uni-giessen.de/?id=3c04b839-a38f-4a9a-bf56-ae8c16ab55ea#_q=bWFrcm8gPSAicyIgLiAvLj8vIC4gbWFrcm8gPSJrZyIgLiBtYWtybyA9Im5pIgomIG9ydGhzYXR6 6ID0gI lMiCiYgIzUgX2lflCMx&ql=aql&_c=R2llc0thTmVfdjAuMw&cl=5&cr=5&s=40&l=10&m=44; Koralek, Ottilie (1889-1890): Lamentatio intermissa I. Tagebucharchiv Emmendingen. Unveröffentlichte Transkription (Hollmann), S. [35] und [43]-[76].

[45] KG = units of cohesion,  NI = non-sentences.

[46] GiesKaNe: https://annis.germanistik.uni-giessen.de/?id=25362693-029a-4e86-962c-198292c378c1#_q=IldlaWwiLiJlcyI&ql=aql&_c=R2llc0thTmVfdjAuMw&cl=50&cr=50&s=0&l=10&m=0; Gessner, Christian Friedrich (1740): Die so nöthig als nützliche Buchdruckerkunst und Schriftgießerey. Bd. 1. Leipzig: Geßner, [0013]-[0019], S. 1-65. Quelle: Deutsches Textarchiv https://www.deutschestextarchiv.de/book/view/gessner_buchdruckerkunst01_1740?p=49.

### 2.1.3 GiesKaNe's grammar model compared to possible de facto standards: From sentence to token in 3 topics (verbs and/or predicates, edge label granularity, interaction treebank and part-of-speech tagging)

Now that tokenization and sentence boundary recognition have been discussed for GiesKaNe (especially given its character as a historical corpus), I now come to the actual core of the syntactic grammar model and the further comparison with possible de facto standards as well as the problems discussed in the discourse. To do this, I will highlight three central themes.

The verb occupies a central position in the sentence, and not only in grammars based on valence theory. Given that TIGER is based on NeGra (Skut et al., 1997), Rehbein (2010: 233) emphasizes that the predicate argument is a central concept of annotation by attaching related arguments to the same parent node. She also focuses on the flatter structure of the trees in TIGER.[47] But in my opinion, TIGER also adopts another concept from NeGra that can theoretically raise questions and whose practical use must be questioned. TIGER (Albert et al., 2003: 48 f.) focuses on the finite verb, which, according to their approach, is fundamental to the analysis of the unit as a sentence (S). In relation to the finite verb, the infinite part is analyzed as a verbal argument of the finite and gets hence the function label OC (object clause) and the category label VP (verbal phrase). 'The subject is always annotated as a dependent of the finite verb.' (Albert et al., 2003: 49)[48] Here, too, there would be a connection point to the previously discussed problem regarding the concept of sentence. But the following is about the verb phrase. An analysis of the TIGER treebank shows that VP is not limited to the combination with the finite verb or tied to the edge label OC (object clausal). However, this constellation can be found in 66.7% of the approximately 24,500 sentences. Another 17% goes to more complex VPs or coordination. In only around 14% of cases the VP does not meet a head (HD) as a clausal object (OC) or (rarely) predicative (PD) – in 3.6% of cases also a modifier (MO). Furthermore, the edge label OC is found 402 times within a noun phrase. Accordingly, it is difficult to integrate the level of VP into the query. A total of 50 combinations of a function and a parent node can be distinguished for VP nodes. For sentences (S) one can distinguish 66 combinations. It is clear that 39 of these pairs show distinct value ratios, revealing significant patterns in their distribution – for example: S-PD: 12 S vs. 889 VP; S-SB: 834 S vs. 72 VP. In contrast, some combinations of functions and nodes are not modularly combinable but are already determined by the subsequent S or VP nodes – for example, in relative clauses (RC) or coordinated VPs (CVP). GiesKaNe follows Eisenberg (2020: 98), who argues for German that analysis with VP does not adequately capture the syntactic relationships. As shown, the distinction between S and VP can also be questioned with regard to the effort and benefit in annotation and query.

A further decision with far-reaching consequences concerns the somewhat more nuanced distinction between predicate and verb. Ágels GTA and GiesKaNe[49] focus on the predicate as the center of the sentence and not (only) the verb. This has the consequence that predicative expressions are analyzed as part of the predicate itself (see Example 4). This view is consistent from the perspective of sentence valence and has consequences for further analysis. With a predicate concept in which the predicate can contain a nominal component (noun or adjective), it should be possible that the complements of the nominal component can also be analyzed with regard to their function in the sentence (GiesKaNe, 2023: 38 f.). The prepositional object "an neuen und wichtigen Actenstücken", which could also be freely moved in a verb-second clause (permutation test) in the sentence and is therefore likely a complement to the sentence, is only required as a prepositional object because of the addition of "reich" and its valence to the predicate.

> **Example 4 (Ranke)[50]:**
> […] **wird** sie <u>an neuen und wichtigen Actenstücken</u> so **reich** , daß sie die Aufmerksamkeit in hohem Grade fesselt .
> <u>Engl.:</u>
> she **is** so **rich** <u>in new and important files</u> ,             that she captivates the attention to a high degree.

---

[47] She compares Tiger and TüBa-D/Z with a view to automatic analysis and she states that this practice leads to crossing edges in the case of non-local dependencies and thus violates the requirements of a context-free grammar. TüBa-DZ solves this task with edge labels that specify the grammatical function (see also Dipper & Kübler, 2017: 606). This leads to flatter trees in TIGER and ultimately has an impact on probabilistic parsing.

[48] In this respect, Rehbein's comparative presentation (2010: 231) of German and English treebank traditions does not seem to correspond to established practice: for the two sentences <u>Der Mann</u> (formal nominative = subject) *beißt den Hund* and <u>Den Mann</u> (formal accusative = direct object) *beißt der Hund* (Engl: 'The man bites the dog' or 'The dog bites the man') she analyzes the subject and the accusative/direct object in sentence two as a child of the sentence node, i.e. always the element in the prefield ('Vorfeld'). In TIGER, however, the subject is linked to the finite verb by rule. The deviation in the analysis underlines that certain analytical traditions appear at least counterintuitive when applied to German.

[49] https://gieskane.com/wp-content/uploads/2023/01/annotationsrichtlinien-1.pdf

[50] GiesKaNe: https://annis.germanistik.uni-giessen.de/?id=88df9fd1-a024-4e25-8db1-f5393e8de110#_q=bm9ybSA9ICJBa3RlbnN0w7xja2VuIg&ql=aql&_c=R2llc0thTmVfdjAuMw&cl=25&cr=20&s=0&l=10&m=0; Ranke, Leopold von (1839): Deutsche Geschichte im Zeitalter der Reformation. Bd. 1. Berlin: Duncker und Humblot, S. III-X, 284-322 und 434-447. Source: Deutsches Textarchiv, https://www.deutschestextarchiv.de/book/view/ranke_reformation01_1839?p=11.

In TIGER (Albert et al., 2003: 64), for example, a complex predicative is assumed for *Er ist [stolz auf Hans]* (Engl.: He is [proud of Hans]). However, TIGER offers a different interpretation in another example for *das Warten* (accusative object) and *leid* (predicative) in the sentence *Sie ist das Warten leid* (Engl., only similar in content, not syntactically: She is tired of waiting) (Albert et al., 2003: 100). This suggests that TIGER may not provide a fully consistent or systematic framework for the analysis of predicates. TüBa-DZ chooses a middle path here and sees an independent prepositional object in such cases (Telljohann et al., 2017: 141), but basically uses its own predicative node (Telljohann et al., 2017: 91 ff.) in addition to the one otherwise used "Verbkomplex" (VC). This issue regarding the boundaries of the analysis of the predicative is by no means a marginal phenomenon of a treebank; rather, it also affects the much more commonly used PoS tagging. Dipper et al. (2013) point out the relevance of a more precise differentiation of the flatter annotation according to the STTS. The STTS provides the tag ADJD for adjectives. HiTS as an extension of the STTS for historical texts (Dipper et al., 2013) extends the usage of D for predicatives. HiNTS for the ReN corpus, by contrast, does not generalize like HiTS (cf. Barteld et al., 2021). In the area of predicates, functional verb structures and verbal idioms are another challenging area of annotation and are documented in more detail for GiesKaNe (2023: 35 ff.). In TIGER (Albert et al., 2003: 63) the basic rule is not to use the label when in doubt. Demske (2007: 100 f.) points out the special challenges with historical texts and, for example, the relevance of lists and statistics.

In the area of the syntactic functions of the sentences, there are less noticeable differences to both treebanks. But, with reference to this syntactic level, another possible de facto standard that is becoming increasingly established can be addressed: Dipper et al. (2024) propose an extension for the Universal Dependencies[51], which they can apply to Middle High German texts and provide good IAA for almost 29,000 tokens. They see inconsistent terminology as the main obstacle to using the existing UD-DE annotation scheme. This runs counter to the German-language tradition in the sense of the two annotation schemes TIGER and TüBa-D/Z that have already been introduced and shows a lack of precision: "some very general labels like xcomp, which covers quite different kinds of phrases" (Dipper et al., 2024 : 17102 f.). I take up this argument and continue the comparison with reference to GiesKaNe. Since GiesKaNe is based on an entire grammar model for practical text analysis, there are some finer categories and labels compared to TüBa-D/Z and TIGER. The syntactic function of the modifier or the corresponding edge label can first be accessed from the form side, i.e., via the child node. In GiesKaNe, example 5a is a free predicative (2,140 pMillionToken)[52], 5b is a temporal adverbial, 5c is a 'Bereichsglied' (scope function)[53] with a relative frequency of 532 pMToken, 5d is a 'Geltungsglied' (validity function, 3,250 pMToken) and 5e is a 'Wertungsglied' (comment function, 833 pMToken). The last three functions can be summarized under primarily semantic differentiation criteria and with an evaluative function.

Free predicatives refer to a single object or the subject: They relate to participants in the scenario (Ágel, 2017: 852, among others). This distinguishes them from adverbials, which primarily reference the sentence or scenario as a whole. In TüBa-D/Z they are all annotated as V-MOD/ADJX and in TIGER as MO/ADJD.

> **Example 5a:**
> (TüBa-D/Zv11, sentence 142): Betrunken ist der miese Sex besser auszuhalten […]
> Engl.: Drunk, lousy sex is easier to endure.
> (TIGER, 8617): Curvat lief seinen Jugendfreunden zufolge unbewaffnet zu den Hunden […]
> Engl.: According to his childhood friends, Curvat ran to the dogs unarmed.
> **Example 5b**
> (TüBa-D/Zv11, 171): […] und prompt wird da gepöbelt.
> Engl: And right away, people start hurling insults.
> (TIGER, 37212): Dort ticken die Uhren schneller […]
> Engl.: The clocks tick faster there.
> **Example 5c**
> (TüBa-D/Zv11, 44): Wirtschaftspolitisch mache ein Auto-Umschlagsort Vegesack keinen Sinn: […]
> Engl: From an economic policy perspective, an automotive transshipment point in Vegesack makes no sense.
> (TIGER, 24289) Zugleich wird beteuert, daß […] ökologisch sowieso alles gänzlich unbedenklich […] sei.
> Engl.: At the same time, it is asserted that, from an ecological standpoint, everything is completely harmless anyway.
> **Example 5d**
> (TüBa-D/Zv11, 44): […] und muß ich nicht an dieser Kontrolle notwendig beteiligt werden […]
> Engl.: And must I necessarily be involved in this control?

---

[51] Existing treebanks from this area are Völker et al. (2019/2023), Petrov et al. (2015), Salomoni (2017), Uszkoreit et al. (2019), Rehbein et al. (2019) (cf. Barteld et al., 2024).

[52] https://annis.germanistik.uni-giessen.de/?id=8f625047-d792-4ec5-932e-01006dfe3bce#_q=bm9kZSAmIG5vZGUKJiAjMSA-W3RyZWU9ImZycHLDpCJdICMy&ql=aql&_c=R2llc0thTmVfdjAuMw&cl=25&cr=20&s=0&l=10&

[53] Here 'ber' in 'kom_ber' can be replaced by 'gelt' or 'wert': https://annis.germanistik.uni-giessen.de/#_q=bm9kZSAmIG5vZGUKJiAjMSA-W3RyZWU9ImtvbV9iZXIiXSAjMg&ql=aql&_c=R2llc0thTmVfdjAuMw&cl=5&cr=5&s=0&l=10

(TIGER, 48340)<u>Wahrscheinlich</u> ist diese Weltgegend das Göttliche schlechthin.
Engl: This region of the world is <u>probably</u> the divine itself.
**Example 5e**
(TüBa-D/Zv11, 306): […] Wer kein Blut sehen kann, sollte bei den dokumentarischen Szenen einer Herztransplantation <u>besser</u> wegsehen.
Engl.: Anyone who cannot stand the sight of blood should (<u>better</u>) look away during the documentary scenes of a heart transplant.
(TIGER, 50463) <u>Natürlich</u> muß sich Rußland ausgegrenzt vorkommen.
Engl: <u>Of course</u>, Russia must feel excluded.

The same applies to prepositional phrases in TIGER, which is illustrated by Examples 6a and 6b: Annotated as MO/PP, they are a part of a possible paired junction *auf der einen Seite* (Engl.: on the one hand), a circumstance adverbial *bei uns* (Engl.: here/in our case*)* and a potential component of the predicate *zum Teufel (gehen)* (Engl.: (go) to hell/the devil) in 6a as well as a circumstance adverbial[54] *bei der Gelben Post* (Engl.: at the yellow post), a validity function *nach einem Bericht der Zeitung die Welt* (Engl.: according to a report in the newspaper die Welt ), a dilative adverbial *bis Mitte 1995* (Engl.: until mid-1995), a local adverbial *in der Verwaltung im Westen* (Engl.: in the administration in the West) and a mediating adverbial *durch die Ausnutzung der Fluktuation* (Engl.: by exploiting fluctuation) in 6b.

> **Example 6a** (TIGER, 1004): […] <u>auf der anderen Seite</u> <u>bei uns</u> die Arbeitsplätze <u>zum Teufel</u> gehen, weil […]
> Engl.: […]on the other hand, jobs are going down the drain here because […]

> **Example 6b** (TIGER, 39337): <u>Bei der Gelben Post</u> sollen <u>nach einem Bericht der Zeitung die Welt</u> <u>bis Mitte 1995</u> <u>in der Verwaltung im Westen</u> 10 500 Stellen <u>durch die Ausnutzung der Fluktuation</u> abgebaut werden.
> Engl. (position of the constituents has not been adjusted): According to a report in the newspaper Die Welt, the Yellow Post plans to cut 10,500 administrative jobs in the West by mid-1995 through natural attrition.

What is striking, as mentioned, is the relevance of semantic aspects of the analysis, i.e., aspects that are detached from the text surface. The group of adverbs[55] including connectors, which are semantically subcategorized in GiesKaNe, also fits in with this. Additionally GiesKaNe distinguishes more clearly than TIGER between syntactic structures at the sentence and phrase level. For practical application in corpus linguistics, the following generally applies: If the labels in the treebank indicate that a grammatical function is implemented within a sentence or word group, this does not need to be reconstructed; If one theoretically assumes uniform functions on both levels, this uniformity must be reconstructed if different function labels are used.[56] The scope of application of the TIGER label modifier (MO) concerns the two levels that are differentiated in the German grammar tradition: 'Satzglieder' (functions in sentences) and 'Gliedteile' (functions in phrases).[57]

Example group 7 shows prepositional phrases within the adjective phrase in TIGER. In addition to the already known polyfunctionality of the prepositional phrase, the spectrum of possible formation patterns for adjective phrases also becomes clear, which can also be differentiated more precisely. In GiesKaNe, attributes are analyzed formally and not functionally at the word group level, while at least a distinction is still made between adjectives, present participles (Participle I), and past participles (Participle II). [58] So the distinction is not made as precisely as the examples require. However, the difference between levels becomes clearer.

---

[54] Ágels GTA (2017) and GiesKaNe analyzes syntactic functions in constituents like this sentence under the premise that each syntactic function can only occur once, and each syntactic function is only fulfilled or needed once. On the one hand, theoretically it would be problematic for "At the Yellow Post" and "in the administration in the West" to assume a single local adverbial, even though a supposed part stands alone in advance. In German, this is the central criterion for accepting a constituent of the sentence. An attribute in a distance position does not do justice to most grammatical theories or the contextual idea of corpus linguistics. On the other hand, the basic rule of not assigning functions in a constituent multiple times seems sensible from the perspective of systematic grammatical description. Section 2.2.4 on annotations in a spreadsheet format shows how this theoretical basic rule is used to generate a unique ID for each constituent during data processing.
[55] https://annis.germanistik.uni-giessen.de/#_q=bm9kZSAmIHRyZWU9InBnciIKJiAjMSA-W3RyZWU9L2Fkdl8uKi9dICMy&ql=aql&_c=R2llc0thTmVfdjAuMw&cl=5&cr=5&s=0&l=10
[56] The labeling of sentences and subordinate clauses in comparison to TüBa-D/Z was already mentioned with reference to the use of hierarchical labels in Section 2.1.2.
[57] Forsgren (2010: 671 f.) compares the terms. In terms of historical development, he makes a reference to Karl Ferdinand Bekker, who classifies the terms in such a way that the equation of functions in sentences ('Satzglieder') and functions in phrases ('Gliedteile') is due to the lack of consideration of the phrases in the structure of the sentence. The comparison is used in the basic structures ('Grundzüge') of a German grammar (Heidolph et al., 1981), in Sommerfeld et al. (1998), in Hentschel and Weydt (2013), Hentschel (2011: 41), Schierholz and Uzonyi (2022) or in Schmid's Old High German grammar (2004/2023). The comparison loses its precision due to the practice of Duden (2009: 800; cf. Hentschel & Weydt, 2013: 357), which applies the term attribute to nouns and thus uses it alongside the term 'Gliedteil'. Furthermore, the pair of terms or the distinction between the two levels often serves as a premise, which becomes clear when analyzing correlates (Wegener 2013a: 415, also 2013b, Pittner, 2013: 452; Breindl, 2013; Holler, 2013: 529; more detailed Zitterbart, 2013 and Ágel, 2017).

> **Example 7** (TIGER):
> **7a** zum Finanzminister bestellte (sentence 56, Engl.: appointed as Finance Minister), **7b** aus anderen Orten herbeigeeilten (sentence 357, Engl.: rushed from other places), **7c** vergleichbar mit der heutigen Situation (sentence 108, Engl.: comparable to the current situation), **7d** für die Studie zuständige (sentence 48993, Engl.: responsible for the study), **7e** um etliches geringer (sentence 1600, Engl.: considerably lower), **7f** Unweit von hier (sentence 318, Engl.: not far from here), **7g** zum Dschungel gewordenen (sentence 3355, Engl.: turned into a jungle), **7h** in der Hinterhand gehaltenen (sentence 2043, Engl.: held in reserve), **7i** im Grund unwichtige (sentence 124, Engl.: essentially unimportant), **7j** zum Teil willkürlich (sentence 506, Engl.: partially arbitrary), **7k** für mich ideal (sentence 840, Engl.: ideal for me), **7l** Frei nach dem Motto des Revolutionsführers (sentence 472, Engl.: loosely following the motto of the revolutionary leader), **7m** fehl am Platze (sentence 2022, Engl.: out of place), **7n** bei weitem nicht ausreichend (sentence 3090, Engl.: by far not sufficient), **7o** vor allem einzelwirtschaftlichen (sentence 14894, Engl.: primarily individual economic)

If you also include the parent node, it becomes clear that the existing group is expanding. TIGER contains almost 115,000 MO relations, which occur primarily in S and VP (76%), but also in adjective (11.5%), prepositional (5%), nominal (5%) and adverb phrases (2.5%). Especially with the modifier edge label used in TüBa-D/Z and TIGER, an already heterogeneous group of contexts at the sentence level is expanded to include others at the phrase level.[59]

This aspect, the granularity of the treebank's labels, is particularly important against the background of effort and benefit as well as accuracy. When discussing granularity, Van den Bosch (2008: 859) also mentions the relationship to the complexity of the source language, Weisser (2022: 91) and Zufferey (2020: 172) draw references to use. It is clear that TIGER's MO cannot be evaluated in this dimension, but here the extremes become clear: Too much granularity reduces accuracy and uniformity, while too little limits usability for specific purposes. GiesKaNe serves as a reference corpus for the syntax of New High German. Corpus linguistics usually tends towards a minimalist orientation on the surface, and it seems sensible to leave further interpretation in the area of the use of the resource instead of integrating it into the corpus resource. As discussed in Emmrich and Hennig (2022), a more rather than less approach generally seems to make sense when it comes to annotations. This not only affects the project's internal research interest in GiesKaNe, but also the fundamental perspective for corpora as a possibility for analyzes of very different questions and research directions in the sense of a crowd/community-sourced corpus. The added value of the complex manual annotations can be seen in merging and comparing. In addition, extensive syntactic analyzes can also be seen as the basis for questions on a 'higher' level – such as research into language, text types, discourses, registers, language change.

As shown, the granularity of the labels and tags can be evaluated in very different dimensions. The key challenge is to establish an appropriate system of granularity. This leads us to the final topic of this section: the relationship between treebank labels and PoS tags.

For PoS tagging in German, the STTS (Schiller et al., 1999) represents a de facto standard. In comparison, HiTS (Dipper et al., 2013) is an adaptation to historical texts and also tends to establish itself as a de facto standard. Emmrich and Hennig (2022: 212) use the example of the adjective to illustrate that even individual tagsets of projects within a larger network of corpora that use this common tagset (HiTS) have been modified.[60] This can also be related to the topic of external (interoperability and usability) and internal project interests. For GiesKaNe, the internal question is how the PoS level can interact with the treebank. With reference to TIGER (Albert et al., 2003: 9), it was already mentioned that the edges in the nominal phrase are not used exhaustively, but the value NK is assigned so that possible syntactic functions have to be reconstructed via the PoS-tags (Section 2.1.3). GiesKaNe relies on exhaustive use of annotation ('more instead of less', cf. Emmrich & Hennig, 2023/2022) and edge tags that define syntactic functions. Accordingly, the PoS tags leave room for further information. We give greater weight to the language system than to the context. Instead of confirming the existing annotations of the treebank using tags (consecutive annotation)[61], information is supplemented if necessary (additive annotation). For example, with the syntactic function of the focus particles – ‚Fokusglied'

---

[58] This corresponds to established practice. This practice is critically discussed in Ágel (2017). The main aim here is to show the possibilities for further differentiation.
[59] On the attribute term: Ágel (2017: 749 ff.), further on the terms complement and supplement: Hölzner (2007), modification: Zifonun (2010, 2016), on the classification between the levels: Welke (2011: 250 ff.) and Knobloch (2015), and Fuhrhop & Thieroff (2005: 315 ff.).
[60] TIGER (Albert et al., 2003: 123) also makes small changes to the STTS (Schiller et al., 1997).
[61] When Dash (2021: 28) emphasizes the interdependence of the levels and states: "POS-level annotation becomes necessary inputs for syntactic-level annotation" the question arises as to whether he does not have more automatic processing steps in mind here.

(Engl.: focusing modifier) in GiesKaNe[62] – we annotate the lexeme classes that are used in this function (source/donor PoS tag).[63] A comparable strategy can also be found in HiTS (Dipper et al., 2013: 92). There, parts of speech are annotated twice – with reference to the token and the lemma. With all of these considerations, the PoS levels must of course represent an internally consistent system that can also be viewed as an alternative access from the aspect of usability. Since this section primarily deals with de facto standards critically, it should finally be clearly emphasized that the advantages of standards have not been discussed in detail here simply because they should be obvious – best expressed by the terms interoperability and usability. But it should be shown that these advantages come with problems. And since it has hopefully become clear that both approaches implement legitimate interests, I would like to finally present a strategy with reference to Emmrich and Hennig (2022) that replaces the either/or situation with a double strategy. Here, HiTS is derived from the existing manual annotations in GiesKaNe.

## 2.1.4 The derivation of a standard from existing annotations and the attempt to do justice to multiple interests

The relevance of such a conversion can also be derived more generally from the framework of Universal Dependencies, which is described, for example, in Marneffe et al. (2021). Among other things, it provides tagset conversion tables – for example from the German STTS to others.[64] The term 'treebank conversion' usually refers to a conversion between constituent structures and dependency structures (Bosco, 2007; Candito et al., 2010, Lee & Wang, 2016), but also between different grammatical structures and labels and there are more general approaches (Wang et al., 1994; Frank, 2001; Sasaki et al., 2003; Han et al., 2013; Lynn et al., 2014; Romary et al., 2015; Çöltekin et al., 2017, Arnardóttir et al., 2020). In the GiesKaNe project, corresponding derivations become important, especially in the final project phase. Derived from the complex annotations, interoperability and usability can be significantly increased. Here we can build on some previous approaches.

Zeman's approach (2008) arises in a situation in which individual transfers are made between two tagsets with rather small differences. Zeman's idea is to make the transfer code reusable via a set of universal features by creating so-called tagset drivers. Basically, he states that tagset conversion is always an "information-losing process", but still worthwhile. Previous approaches primarily pursue the goal of tagset standardization (EAGLES: Leech & Wilson, 1999; LE-PAROLE: Volz & Lenz, 1996; Bacelar et al., 1997; MULTEXT: Ide & Véronis, 1994; MULTEXT-EAST: Erjavec , 2004; cf. Zeman, 2008). Compared to Zeman's rule-based approach, the approach of Zaborowski and Przepiórkowski (2012) is statistical and uses decision tree classifiers. They claim a precision of 94.95% and note that they achieve "significantly higher accuracy than state-of-the-art taggers". More generally, they conclude that "tagset conversions – in the sense of re-tagging a corpus – can be significantly improved by using information taken from the previous manual annotation." The fact that the token is still the central resource in this process should come as no surprise when taggers can achieve over 97% (Manning, 2011) accuracy under 'optimal'[65] conditions. The authors also ultimately point out that the approach shows its advantages when very good results are achieved with little training data. The central point on which the present program is based.

In Emmrich and Hennig (2022), 4 GiesKaNe texts (4 x 6,000 = 24,000 tokens) are supplemented with an annotation according to HiTS. In order to obtain training data, it might be better to annotate texts from corpora that already use annotations according to the target tagset. Because then the manual annotations are always carried out by the experts for the respective tagset. Emmrich and Hennig (2022) compare a model for automatic PoS tagging with manually annotated features versus one based solely on features of the token in the context. The model based on the derivation from the manual annotations achieves an accuracy of 96% after being trained on a single text containing 6,000 tokens, compared to the 'simpler' model (CRF tagger utilizing token-level character sequences and contextual features), which achieves an accuracy of approximately 85%. Additionally, accuracy of the next text's first token segment does not fall below 92.5% and then increases continuously, while the simple model repeatedly shows drops to around 77%. In practice, the study was implemented using a Conditional Random Field/CRF (Lafferty et al., 2001). A preliminary work from 'normal' PoS tagging is Zhu et al. (2013), who achieve an accuracy of 96.6%. Since Emmrich and Hennig (2022) achieve their results with historical texts (some conceptual oral texts) based on small quantities of training data, the comparison to historical taggers is more appropriate. Dipper (2010, 2011) achieved the best results with 92.9% for Middle High German texts that were normalized and annotated with a precursor to the later HiTS tagset[66] and for which the

---

training data fit the test data very well. For comparison, in the same work, an accuracy of 95.7% was achieved with 210,000 tokens and 94.4% with 90,000 tokens when applied to contemporary German or TIGER. For Middle Low German texts, Koleva (2017) reported an in-domain accuracy of up to 87.7% and achieved 77.1% accuracy in a cross-genre robustness experiment on a religious prose dataset. As mentioned, the advantage of deriving from existing (manual) annotations is that, as abstractions over contexts, they are not or at least hardly subject to these influences. Schmid (2019) uses a bidirectional LSTMs with character-based word representations and can present an accuracy of 98.23 for the GerManC corpus (Durrell et al., 2007; newspaper texts, 1650-1800) and 95.9% for REM (Petran et al., 2016). The values are difficult to compare: In GerManC the STTS (54 tags) with 9 modified tags (Durrell et al., 2012: 9f.), which make up 2% of all tokens in the Gold Standard sub corpus. HiTS has 72 tags. Newspaper texts from New High German are certainly less dependent on normalized word forms than different types of text from the Middle High German or show less orthographic systematics. The proportion of orthographic symbols within the sentence is also crucial in this area, as Manning (2011) points out: 'because you get points for every punctuation mark and other tokens that are not ambiguous'. In this respect, the already considerable value for Middle High German seems to be the better reference point. For an exact comparison, we train Schmid's RNN tagger[67] with 80% of the sentences (without a final punctuation mark) from each of the 24 GiesKaNe texts (v. 0.3) and test the result with the remaining 20%. Conceptual oral and literal texts are mixed; But there is already training data for each test text, so that text-specific requirements have been taken into account – more than one can expect in practice. In practice, the values could be even lower. We achieve an accuracy of around 92% in the test data set, which can be a considerable good value if you take into account the strong orthographic variation between the texts, a lower accuracy score by fewer orthographic sentence boundary marks and an extended tagset that integrates syntactic-semantic information. The entire range of values also seems acceptable when compared with IAA studies: STTS/NEGRA (contemporary language newspaper texts): 98.57% (Brants, 2000), HiNTS/ReN (Middle Low German/Lower Rhine): 94.33% (Barteld et al., 2018), STTS-EMG/GerManC (New High German): 91.6% (Scheible et al., 2011). For the part of speech level in GiesKaNe and our tagset, the value is 95.8% (see Section 2.2.3.2). Schmid's (2019) 'alternative' is also interesting in comparison to the GiesKaNe workflow with its alternation of human and automatic work steps (Section 2.2.3). Here, with the article by Schulze and Ketschik (2019), I would like to address the role of the scope of the training data. Because Schulze and Ketschik (2019) report significantly worse values than Schmid (2019) and Emmrich and Hennig (2022) for their experiments on Middle High German with a CRF and a comparable token scope as well as a significantly smaller tagset that is oriented towards the UD (Nivre et al. 2016): With 6,000 tokens, the accuracy slowly goes above 80%, then only slowly reaches a value of 87%. On the one hand, this speaks in favor of Schmid's (2019) approach, and, on the other hand, it shows that a very high level of accuracy can be achieved relatively quickly by using existing annotations. This is important because parallel annotation (e.g., according to multiple annotation schemes, tagsets, or a de facto standard) is an investment that has to be justified during corpus compilation but seems sensible in terms of the cost-benefit ratio.

## 2.2 The workflow: The alternation of human and automatic processing steps, text normalization, parser evaluation, IAA and the construction of a treebank in a spreadsheet

### 2.2.1 Machine normalization: goals, approaches and practical application

I begin with the question of the purpose of the normalization process, leading into the practical application of automatic normalization and concluding with a study on its accuracy within GiesKaNe. As with tokenization, the basis of this step is GiesKaNe's guidelines for text preparation ('Richtlinien für die Textvorbereitung'). Essentially, normalization involves orthographic adjustment, which inherently brings phonetic adaptation to the standard pronunciation. Morphological adjustment, on the other hand, would be a significant intervention in the grammatical core structure of historical texts. With this juxtaposition, the field of normalization is initially outlined in such a way that broader acceptance can be assumed. However, this overview remains too open, requiring the underlying problems to be addressed specifically. As Odebrecht et al. (2017: 704) aptly summarize, normalization in corpora can serve different purposes: 'finding instances of the same word, making generalizations, enabling linguistic processing'. Similarly, in the DTA, it is stated that orthographically inconsistent texts should be searchable in such a way that graphemic variants of a word (e.g., *Kleid*: *Kleidt*, *Kleydt*, *Cleyd*, *Cleit*, etc.) can be found with a single query[68], while punctuation is not normalized to modern standards.[69] Quotation marks around 'same word' in Odebrecht et al. (2017) likely indicate that there should be more problematic examples than the example of *Kleid*. Differences also emerge between normalization for a

---

specific research question and normalization for automatic processing: the former seeks variation, the latter uniformity. Odebrecht et al. (2017: 704 ff.) describe, for example, two normalization levels for the RIDGES corpus (Lüdeling et al., 2020), which they call CLEAN and NORM. CLEAN is generated automatically, requiring that all letters be replaced by an equivalent in contemporary German. This also applies to umlauts (*ä, ö, ü*). The DTA states that the handling of diacritics should be normalized if they can be mapped to present-day language.[70] This step, for example, is carried out in GiesKaNe at the token level, as documented. Although this may seem like an intervention in the text surface, it is justifiable, especially in comparison with RIDGES and the DTA, as the consequences show. Tokens represent a fundamental level of the corpus, not necessarily a reflection of the text surface. They mediate between surface and analysis, between accuracy and usability. Adjusting tokens in GiesKaNe is feasible because the reference to the text surface is secured through the simple segmentation of DTA tokens. Such a change in the tokens compared to the text surface – beyond white space tokenization – is possible in GiesKaNe because this surface reference is retained via the DTA token level, and this level is mapped onto the tokens. The DTA tokens, in turn, ensure the connection to the TEI standard and to the scans of the original texts. But the mediating role of tokens then has an impact on aspects of usability: the treebank in ANNIS is based solely on the tokens. The normalization must be integrated in addition to the already quite complex query of syntactic structures (both are only linked via the token level). In this respect, we adapt characters to modern spelling at the token level (to make querying easier), but do not make any 'changes' beyond that. RIDGES, in turn, requires two normalization levels – NORM and CLEAN. In this context, the authors also mention that the goal of this normalization step does not necessarily have to be contemporary German; it could also be a historical language stage, and assigning historical spellings to modern word forms always involves interpretation.[71] The DTA goes one step further; virtually works without a normalization level.[72] However, this means that a special reference to the token level or the text surface has to be made when querying. The query *Kleyd*[73] stands for the lexeme and in *"@ich es"*[74] this behavior is switched off with the @ operator for *ich*, while *es* also includes *ʼs*.

In line with the discussed top-down syntax of the GTA, the annotation workflow in the GiesKaNe project begins with the manual breakdown of the texts into, among other units of the texts, grammatical sentences (for a definition see Section 1.2). This is preceded by a text selection process (cf. Section 2.3) and a machine preparation phase. A spreadsheet is created for the annotations, where cells are merged, and values are entered. The rule-based tokenization process in GiesKaNe was already addressed (Section 2.1.1), showing that challenges in automatic tokenization arise primarily because it is combined in one step with automatic normalization. This integration is done to streamline a manual step for tokenization and normalization, or their verification, with manual text segmentation. For this reason, we avoid using specialized tools[75] for manual normalization, such as VARD 2 (Baron & Rayson, 2008), Norma (Bollmann et al., 2012), or Orthonormal (Schmidt, 2012/2016).[76]

However, the best practical application is central to our corpus compilation. The basis for this so-called GiesKaNe basic version of a GiesKaNe text to be annotated in the form of a spreadsheet (see Fig. 1) is usually a text from the DTA. This applies especially to conceptually written texts (see Section 2.3). The web service CAB[77] (Jurish, 2012) with simple whitespace tokenization and 'suggestions' for normalization forms the basis. The normalization is based on "a robust generative finite-state canonicalization architecture" (Jurish & Ast, 2015; cf. Jurish, 2012) via Hidden Markov Model. Jurish (2012) compares different methods of normalization: string identity (id), transliteration (xlit), phonetization (pho), rewrite transduction (rw), hidden Markov model (hmm): The Hidden Markov Model "performed best at the token level, achieving a token-wise harmonic precision-recall average F = 99.4%" (Jurish, 2012: 65) with a precision of 99.7 and a recall of 99.1. The comparison shows that transliteration, "which employs a simple deterministic transliteration function to replace input characters which do not occur in contemporary orthography with extant equivalents" (Jurish, 2012: 51 f.), can also deliver good results (prec = 99.8; recall = 96.8, f = 98.3). In Jurish and Ast (2015) the HHM is contrasted with the use of a semi-automatically constructed "finite deterministic canonicalization lexicon" and a "a hybrid method which uses a finite lexicon to augment a generative canonicalizer": "The observed results showed that while both the HMM and corpus-based techniques were quite effective on their own, the hybrid technique outperformed both of them in both type- and token-wise F." The following values are obtained for the HMM and the tokens: p = 0.996, r = 0.985, f = 0.985. For the lexicon, f is 0.992 and for the hybrid it is 0.995. Previous approaches – overviews are provided by Baron (2011: 21 ff.) and also Jurish (2012: 2 ff.) – which use letter replacement heuristics and similarity measures, deliver different results (Pilz et al. 2006; Bollmann et al., 2011a, Bollmann et al., 2011a 2011b; Bollmann, 2012). As expected, different texts emerge as relevant for the results. Bollman et al. (2011a) and Bollmann (2012) compare the Luther Bible and manuscripts from the Anselm corpus from the Eastern Upper Germany dialectal area. The basis of the comparison is always a normalized version. A baseline is created for the ratio. They note that the baseline (in the sense of the number of words which do not differ at all between old and modernized versions) is significantly lower for Anselm (35.42%) than for Luther (64.71%). They state: "It turns out that substantial variations in spelling, as they occur in the Anselm texts, are problematic for the rule-based approach, as it cannot normalize character-context sequences that have not been previously learned." (Bollmann et al., 2011a) Accordingly, the measure used also comes into focus. This is quite clear in Bollmann et al., 2011b: "[A]ny normalizing process that results in less than 65% exact matches [(the baseline)] has probably done more harm than good, and it would be better to leave all words unchanged." The study by Bollmann et al. further clarifies. (2011b) that the best rule-based and the best probability-based approaches achieve match ratios above 83% – the best-probability method with the dictionary lookup even yields 91% exact matches. And further: "Our normalization approach is not only successful in changing historical forms to modern ones, but also in correctly leaving most of the wordforms unchanged that do not need to be changed (97.46-99.57)." Compared to this set of established approaches or Jurish (2012), Ehrmanntraut's (2024) *hybrid_textnorm* incorporates a state-of-the-art approach (transformer language models, encoder-decoder model, pre-trained causal language model). As I said, the conditions for a real evaluation cannot and should not be created here, as with CAB. However, a comparison with CAB shows that although the precision is in the same range, the recall drops noticeably. This is particularly evident in the conceptual oral texts such as Briefwechsel (correspondence) and Bauernleben (Chronicle) with values below the values for Jurish (2012) given in Fig. 5.[78] So no other methods than the previous ones are necessary for our texts and Jurish (2012) will be examined in more detail below, because this section also ends with a study based on GiesKaNe.

---

[70] https://www.deutschestextarchiv.de/doku/basisformat/mdDiakritika.html

[71] Especially against this background, the reference to contemporary German does not seem to be the worst choice: the interpretation here at least has a stable basis due to the high degree of standardization.

[72] Without reference to the DTA, Odebrecht et. al (2017, see Dipper, 2015) introduced the term 'fuzzy search' for the "mapping of different forms in the search itself". This would add to the work of Zobel and Dart (1995) and Pilz et al. (2009) fit (cf. Baron, 2011).

[73] https://www.deutschestextarchiv.de/search/ddc/search?fmt=html&corpus=ready&ctx=&q=Kleydt&limit=10

[74] https://www.deutschestextarchiv.de/search?q=%22%40ich+es%22+%23random&in=text

[75] In this list from Clarin, the term 'tool' is used more broadly: https://www.clarin.eu/resource-families/tools-normalization

[76] An interesting but overly broad question would be, given the extensive discourse on methods of automatic normalization being practically addressed here, whether this mutual relationship (consider examples like *soldens*, *folgeſtu*, *lernetens*, *liſſens*, see Section 2.1.1) was accounted for in the evaluation.

[77] https://deutschestextarchiv.de/public/cab/

To do this, all 24 texts from GiesKaNe version 0.3 were analyzed again with CAB[79] (Jurish, 2012): The DTA token layer of the GiesKaNe texts was uploaded. In terms of precision, recall and F-score, the result of the automatic analysis according to Jurish (2012) was compared with the manual one in GiesKaNe. True positives are a change to the DTA token that was necessary (GiesKaNe normalization ≠ DTA token) and that was implemented by CAB (CAB normalization = GiesKaNe normalization). Everything else can be counted/calculated accordingly.[80] In the further analysis, the different pairs were collected and grouped according to their frequency per text and then they were classified accordingly, so that in the end the actually interesting pairs remain. The analysis then shows that the accuracy values discussed so far (in the literature) are not achieved, which is particularly clear from the recall (Fig. 5). Here again there are clear differences between texts such as Anthus and Michelis (0.973, 0.972) on the one hand and Bauernleben and Guentzer (0.439, 0.491) on the other. In anticipation of Section 2.3, the 'Nähe-Distanz'-values (COL-values = conceptual orality and literacy value) values of the GiesKaNe text (Emmrich, 2024, cf. Koch und Oesterreicher, 1985) can be used here, because all texts highlighted in bold are texts with moderate or strong conceptual orality. These morpho-syntactically based values correlate with the orientation towards an orthographic standard or the (more difficult) traceability to such a standard (COL-value (Section 2.3), F-value (Fig. 5): $r_{Pearson}$=0.71, p < 0.001): Conceptually oral texts tend to have lower F-scores in automatic normalization. Of course, the development of an orthographic standard and the mastery of one is in no way tied to the concept of immediacy and distance (Nähe und Distanz, Koch und Oesterreicher, 1985) itself.

---

[78] One reason could be that there are hardly any contexts that come close to standard language usage and can serve as an orientation for the model. In these texts, the (orthographic) sentence is also not a fixed unit, as discussed (Section 2.1.2). Since long input sequences (e.g., sentences) increased the processing time problematically when using hybrid_textnorm, such inputs (e.g., over 30 tokens) were broken down into segments of 15 tokens each. This may also have worsened the model performance.

[79] https://deutschestextarchiv.de/public/cab/, Analyzer: 'all'. Adjustments to 'norm' would have been possible here – for example to the centuries. However, since the system is not intended to be evaluated here and the results clearly show that conceptual orality and writing are more important, no more data series were included.

[80] The fact that the conditions to evaluate the normalization with CAB are not met can be illustrated by two common but for the evaluation not relevant differences: GiesKaNe normalizes to *dass* (Engl.: that) or *ss*, CAB to *daß* (Engl.: that) or *ß* and in GiesKaNe abbreviations like *l.* for *liebe* (Engl.: dear) are resolved. It is clear that CAB does not want to do that. This can once again illustrate how many individual, different decisions contribute to achieving a single clear goal.

| DTA CAB (Jurish, 2012) | anthus | bauernleben | becher | birken | braeker | briefwechsel | dietz | forster | freyberger | gessner | guentzer | harenberg | knigge | koralek | mendelssohn | michelis | nehrlich | nietzsche | mn_weltmann | ranke | simmel | soeldnerleben | thomasius | zimmer |
|---|---|---|---|---|---|---|---|---|---|---|---|---|---|---|---|---|---|---|---|---|---|---|---|---|
| p | 0.996 | 0.95 | 0.993 | 0.988 | 0.984 | 0.982 | **0.911** | 0.985 | 0.986 | 0.995 | 0.957 | 0.979 | 0.995 | **0.928** | 0.999 | 0.995 | 0.986 | 0.973 | 0.997 | 0.988 | 0.983 | 0.996 | 0.997 | 0.99 |
| r | 0.973 | **0.439** | 0.766 | 0.83 | 0.861 | **0.664** | **0.63** | 0.864 | 0.832 | 0.901 | **0.491** | 0.714 | 0.919 | 0.817 | 0.942 | 0.972 | 0.779 | **0.653** | 0.85 | 0.948 | 0.798 | **0.631** | 0.879 | **0.589** |
| F | 0.984 | 0.601 | 0.865 | 0.902 | 0.919 | 0.793 | 0.745 | 0.92 | 0.902 | 0.946 | 0.649 | 0.826 | 0.956 | 0.869 | 0.97 | 0.983 | 0.865 | 0.786 | 0.918 | 0.968 | 0.881 | 0.772 | 0.934 | 0.738 |

Fig. 5, Precision, recall and F-score for normalization with CAB (Jurish, 2012), evaluated based on agreement with manual normalization in GiesKaNe

The trend continues in the absolute numbers of normalizations. A Welsh T-test (as well as the Wilcoxon rank sum test) shows a clear predominance of the conceptually oral texts: Mean$_{conceptual\ orality}$ = 1380 (std = 945), Mean$_{conceptual\ literacy}$ = 492 (std = 353); t = -2.7, df = 9.36, p = 0.023. If the 3 centuries of the GiesKaNe corpus (17th, 18th, 19th centuries) are included as independent variables, this accounts for 80% ($R^2_{adj}$ = 0.8, F = 18.83, p < 0.001) of the variance in the number of normalizations between the texts explained. Not unexpectedly, there is a strong, significant interaction between the expression of conceptual oral text and 17$^{th}$ century: the number of normalizations per text increases by 1161.1 compared to the intercept (p = 0.004). Accordingly, the time for this work step can be weighted accordingly, which can be supported by automatic classification in advance (see also Section 2.3). The type of changes to be made can be discussed using examples, preceded by some practical figures. GiesKaNe v03 includes around 343,000 DTA tokens, of which 293,000[81] putative word forms are considered here. Of these, around 77,600 had to be adjusted. This was done 59,000 times by DTA-CAB (Jurish, 2012) according to the value in GiesKaNe v03, which is also used here as the basis for the comparison. This results in 18,600 false negatives. There are also around 1000 false positives (p = 0.984, r = 0.760, f = 0.857). They are not central, which is why I will only briefly mention the most important points here: Names are often predictably problematic, much may not be known. Although *Goethe* was surprisingly changed to *Götte*, then *Nitzsche*[82], a rope maker's daughter, was changed back to the better-known philosopher. In the case of *Bricken*[83](a small fish)[84], which was incorrectly changed to *Brücken* (Engl.: bridges), text recognition (Anthus, lecture on the art of eating) as well as dialectal aspects or word embeddings[85] or contexts ("Geschmack leckerer Bricken, Forellen, Hechte […]", Engl.: taste of delicious bream, trout, pike […]) could be helpful. Systematically more interesting are tokens like *ward*, which was wrongly changed to *wart* – i.e., 1st/3rd. person, past tense of *werden* (Engl.: become) vs. 2nd person plural. What is interesting for GiesKaNe, and the general discussion is the group of tokens *sahe*, *ihro* oder *ihme*. The GiesKaNe guidelines state that old inflexibles will not be removed from personal pronouns. So again it is about making decisions – also against the background of research interest. The apocope is therefore discussed in many case studies in GiesKaNe's guidelines for text preparation. The false negatives again show this issue of certain definitions, but also that the automatic model does not disambiguate well when two reasonable alternatives are present. Almost 3,400 unmade changes are based on the following pairs:

das vs. dass[86] (457), andern vs. anderen (434), wann vs. wenn (223), den vs. denn (174), wider vs. wieder (135), dan vs. denn (123), vor vs. für (102), wahr vs. war (96), will vs. will (95), dann vs. denn (86), darnach vs. danach (85), wen vs. wenn (82), unsern vs. unseren (80), unserm vs. unserem (80), Man vs. Mann (71), Leut vs. Leute (68), Tag vs. Tage (59), unsre vs. unsere (58), Gelt vs. Geld (53), stehet vs. steht (53), were vs. wäre (50), gehöret vs. gehört (47), dan vs. dann (47), wahren vs. waren (43), Statt vs. Stadt (43), Nager vs. nacher (42), Tier vs. dir (42), gar vs. Jahrs (40), nit vs. nicht (39), Bey vs. Bei (37), gehet vs. geht (37), uber vs. über (36), bestehet vs. besteht (35), hat vs. hatte (34), sol vs. soll (34), unsrer vs. unserer (33), vil vs. viel (33), fur vs. für (33), in vs. ihn (33), Konten vs. konnten (32)

In principle, syntactic-semantic disambiguation is required here, and it becomes clear that an automatic analysis of the syntax can rely less on the tokens of the text's surface. In future applications, Schmid's (2019) RNN-Tagger approach (see Section 2.1) could compensate for possible weaknesses that current approaches such as Ehrmanntraut's (2024) hybrid_textnorm still show. However, this requires a more precise evaluation of both tools for different types of text in the range of conceptual orality and literacy (see Section 2.3). Overall, the pairs show an obvious tendency: correct normalizations range between one and two Damerau/Levenshtein distances (Damerau, 1964; Levenshtein, 1966) with a tendency towards a single change (mean = 1.383, std = 0.660). Incorrect normalizations from the area of relevant elements/changes – i.e., those that need to be changed – are logically well below 1 (mean = 0.574, std = 0.994) and the target norm well above 1 (mean 1.843, std = 1.156) (Fig. 6).

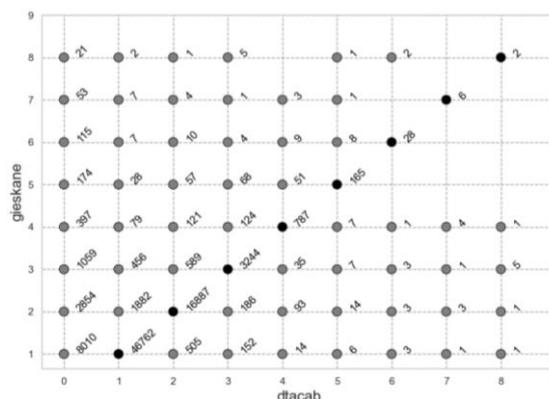

Fig. 6, Damerau/Levenshtein distance between token and DTA-CAB (x) and GiesKaNe as target value (y), only for the cases token ≠ GiesKaNe normalization

The binary logistic regression shows a positive coefficient for CAB (4.58, $z = 141.7$, $p < 0.001$) and a negative one for the GiesKaNe values (-4.81, $z = -121.9$, $p < 0.001$), which ultimately only means that with more necessary transformations the probability of a match decreases and with more attempted transformations of CAB the probability of a match increases. This is basically due to the data set with only relevant cases (modifications to be made), but overall it fits with the lower recall. Fig. 6 (outliers not shown) also shows that the Damerau /Levenshtein distances are within a manageable range for human observation, and they often concern replace operations, which, as shown, are more systematic: *Fꝛantzößiſche* vs. *französische* with 7 Damerau/Levenshtein distances, *taufendfünffhunderteinvnddreyſſig* vs. *tausendfünfhunderteinunddreißig* with 8 or *Mißverſtåndnüſſen* vs. *Missverständnissen* with represent the rare upper end of adaptations and primarily concern direct character exchange.

Ultimately, there is a chicken-and-egg problem for machine analysis if you cannot resort to manual corrections. GiesKaNe relies on sensible preparation of manual correction steps, whereby, in addition to CAB, a normalization lexicon based on previous normalizations in GiesKaNe and the Grimm dictionary are integrated in order to mark the suggestions for normalization in different colors (indicated in Fig. 2). What is crucial is an efficient first manual work step together with the tokenization and segmentation of the text into macro segments (e.g., grammatical sentences). Further improvements could be achieved here against the background of syntactic-semantic disambiguation through characters and/or word embeddings, as well as LLMs.

This analysis of frequent normalization patterns should not obscure the fact that the actual time-intensive work is manual. For GiesKaNe, manual normalization based on the described machine-assisted preparation can be estimated at roughly 2,000 tokens per hour. However, experience reports confirm that the differences between individual texts are a central issue: texts with more conceptually written language are generally described as significantly easier and faster to normalize. As already hinted at, this process begins with semantic disambiguation but goes beyond it. Unlike tasks such as verifying PoS tagging, certain tokens here require considerably more time compared to 'unproblematic' cases, especially when resources like the 'Wörterbuchnetz'[87] (Engl.: dictionary network) are consulted, and more complex cases are discussed. For instance, *anfahen* and *Garte* can be examples where inflection is not altered, but despite their presence in Grimm's dictionary[88], they are lemmatized to user-friendly, easily searchable forms such as *anfangen* (Engl.:

---

begin) and *Garten* (Engl.: garden). With these comments we can now move on to the central theme of the section. Now that the basics of tokenization, normalization, and macro analysis, as well as the central aspects of the grammar model, have been presented, their interaction can be explored in greater depth, and the workflow in GiesKaNe can be outlined. What is crucial is how manual and automated work steps complement each other in a meaningful and efficient way.

### 2.2.2 The workflow: The sensible alternation of mechanical and manual work steps

As mentioned, complementary alternations between machine and manual work steps determine the workflow and the transfer of data between different formats that are more readable by humans (or machines) (Fig. 7): machine steps (gray), manual steps (white) and formats (dark gray). Text selection was not supported by machines until now (v0.3). Section 2.3 introduces an automatic approach.

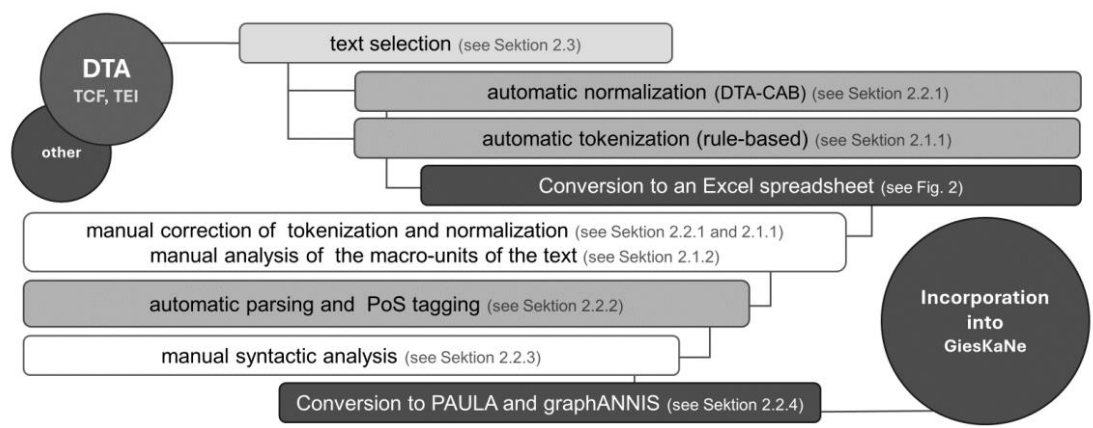

Fig. 7, The workflow of the GiesKaNe corpus: manual and automatic work steps as well as formats

The text selection is based, if possible, on the texts available in the DTA. These have a coarser whitespace tokenization compared to GiesKaNe (corresponding to processing with CAB; Jurish, 2012; see Section 2.1) and there are 'suggestions' for normalization. Texts not included are processed using CAB. The data is in TCF format[89], a clear, multi-level standoff XML format in a single XML document. Practical experience has shown that the TEI format must also be used in order to be able to automatically remove footnotes from the body text. The target format for the transfer is a spreadsheet (e.g., MS Excel) (Fig. 2, Fig. 9), which enables graphical viewing and changes by connecting cells. The advantages: Spreadsheets can be used on any PC, they are stable, MS Excel offers interfaces to extensive Python APIs[90] (underlying information is stored in XML format in the background). They are suitable for collaborative work in the cloud, offer freedom for any additions (including comments), provide graphical support, and allow for the automation of work steps via macros. The use of spreadsheets is unproblematic in practice. However, as outlined in the following text, our ideas for an optimal annotation tool highlight that a spreadsheet should still be regarded as an interim solution.

### 2.2.3 Interactions: Human and machine as well as Human and Human

Syntactic parsers and PoS taggers are central to a project like GiesKaNe. Research on methods is also extensive under the influence of neural networks and numerous models are offered, whereby the optimization usually affects a few percentage points of accuracy in a high range, while the quality when applied to less standard language texts usually remains open. In this situation, a project must find an efficient solution that corresponds to the state of research. Compared to optimization in the comma range, it is again important to see accuracy as work achievement against the background of manual steps. I will start with a discussion of the interaction of machine and manual work steps in comparable projects that look at exactly these two dimensions.

In an early study, Marcus et al. (1993) use tagged and parsed data and correct it. Compared to purely manually created test data, they notice an acceleration of the annotation process and an increase in accuracy and the Inter-Annotator Agreement (IAA). For PoS tagging, values of around 20 minutes are given for 1000 words compared to 44 minutes. This aligns with proportions observed in qualitative observations or case reports regarding work within GiesKane, which I will address at the end of this section. In terms of parsing, Chiou et al. (2001) report an improvement from 240 words per hour to almost 480 words per hour. Here, our results for phrase analysis are

---

positive, but with a time saving of up to 50%, they are not quite as positive. Similarly, Xue et al. (2002) report only a time saving of 42% for parsing compared to purely manual analysis, without compromising quality. Felt et al. (2014) specify that human annotations are improved when 'even mediocre pre-annotations' are provided: "When pre-annotations are at least 70% accurate, annotator speed and accuracy show statistically significant relative improvements of 25-35 and 5-7% , respectively." Eckhoff and Berdicevskis (2016) are also more specific and show in their study of Old East Slavic with the Malt-Parser (0.804 UAS, 0.779 LAS), experienced annotators became between 5.8 and 16.57% faster, and inexperienced annotators between 14.61% and 32.17%. There were no differences in accuracy. In contrast, Rehbein et al. (2012) found no acceleration of annotations in experiments on semantic role labeling, but an increase in quality. In experiments on named entity recognition in clinical trial announcements, Lindgren et al. (2014) found a statistically significant acceleration and no distortion of the annotation quality through pre-annotations.

Mikulova et al. (2022) present extensively evaluated experiments for dependency-parsed structures and note that pre-annotation not only accelerates the annotation process, but also increases the consistency of the annotations without affecting their quality. This distinction seems sensible because, with the persistence of a machine, definitions made in the annotation guidelines could be constantly overwritten – via the suggestions perceived by the human annotators. Fort and Sagot (2010) confirm in their experiments that pre-annotations increase quality in terms of accuracy and reliability in terms of the IAA; But they also state: "We also demonstrated that this comes with biases that should be identified and notified to the annotators, so that they can be extra careful during correction." This is contradicted by the studies by Rehbein et al. (2012: 19), who state, among other things: "What we could show, however, is that even incorrect pre-annotations do not corrupt the human annotators, and that the reliance of human annotation on these instances is demonstrably not worse than on unannotated text." In GiesKaNe, for example, it is noticeable that the syntactic structures of the treebank repeatedly lead to discussions and a comparison of several levels. Thanks to the top-down approach (see Section 2.1.2), the sentence boundaries determined at the beginning are constantly questioned. PoS tagging, on the other hand, forms the other end of the process and is corrected relatively quickly. It is therefore important to build in further controls – such as a second tagger model for comparison (Zinsmeister, 2015: 96).

Above all, however, new models must be trained with new or more extensive data, if possible. Felt et al. (2014) point out that machine learning relies on training data, which would become clear at the beginning of the training process. The quality is only developing slowly and could cause more damage, although they see advantages, especially for PoS tagging. This would also fit with the findings of Fort and Sagot (2010), who state that a helpful pre-annotation system can be built with even a small amount of training data.[91] With reference to Carter (1997), Oepen et al., (2002) and Ringger (2008), Felt et al. (2014) suggest the possibility of offering alternatives. While the approach is undeniably interesting, in my opinion the implementation in a tool would have to be discussed in the light of availability, sustainability, trainability to an annotation scheme and usability. One example is Annotate (Brandts & Skut, 1998), which was widely used but unfortunately has not been maintained for a long time.[92] An attempt was made there to bridge the gap between shallow and full parsing using chunk tagging (see also Manning & Schütze, 1999: 376). Dipper and Kübler (2017: 610) highlight the advantages:

> "Instead of generating the entire sentence structure in one step, the parser only generates one local subtree in each step, which is immediately checked by the human annotator, and modified if necessary. Based on the annotator's decision, the parser generates the next subtree, and so on. The advantage of this kind of interactive parsing is that the automatic parser can use the decisions made by the human annotator at lower levels. In this way, errors from the statistical parser do not propagate to higher levels, and can often be detected more easily since the annotator's focus is always on the node generated most recently."

Similarly, in connection with GATE (Bontcheva et al. 2013) – primarily intended for flat annotations – an ideal workflow is drawn against the background of mixed-initiative development like in Alembic (cf. Day et al., 1997). Our experiences can be stated accordingly: Parser analysis (consequential analysis) that is based directly on a human annotation decision seems ideal, especially for historical texts. Whole sentence analyzes – as in pre-parsing the entire text – are beneficial if they only need to be viewed and confirmed or if changes primarily affect the terminal nodes and thus affect less sibling or child nodes.  If you look at a sentence from the perspective of syntactic depth, errors in higher nodes have much more (negative) effects. And here too, the values differ noticeably for different texts and relative to the sentence length (see Section 2.3) and the number of normalized tokens (Section 2.2.1). The article by Sapp et al. (2023) points in a different direction here: The authors examine the effect of pre-annotations when parsing Early New High German texts with manual correction (annotations in Penn Bracketing Format) and compare two texts of different complexity and syntactic parser analyzes of different depths:

> "The results […] suggest that, when the text is syntactically simple like Karrenritter (mean sentence length of 18.6 words), correcting from the output of the neural parser is no faster than correcting from a minimally parsed text. However, in a syntactically more complex text such as Geistliche Mai (mean length 30.5 words), manual correction is much faster when the text was parsed by a neural parser, either with or without GFs [(grammatical functions)]." (Sapp et al., 2023: 60)

---

[91] This is different, for example, from Meelen and Willis (2022) in the context of the creation of a treebank of the middle and modern Welsh, who use a memory-based tagger (Daelemans et al., 2010) with better results than a neural network-based one (Chernodub et al., 2019) because there is little training data for Middle Welsh.

[92] According to Dipper and Kübler (2017), the tool was used in NEGRA, TIGER, TüBa-D/Z, Verbmobil, Potsdam Commentary Corpus, Mercurius Treebank, Deutsche Diachrone Baumbank and SMULTRON, but is unfortunately no longer maintained.

From the data you can also see that for the syntactically less complex text there are 392 words per hour for the very simple rule-based approach to generating phrases and 352 for the complex parser model, while for the more complex text there are 273 and 352. However, the authors also consider the findings included here for other texts: " […] [T]he parser may just be able to create more efficient trees on the shorter sentences, which are often syntactically simple." In this respect, one could also interpret the available figures for correction time to mean that a short correct sentence is ideal. For long sentences, incorrect analysis could still save time. This would be more in line with our experience.

Based on experience reports and smaller studies, the following observations can be made for the GiesKaNe corpus: In the pre-parsed manual phrase analysis, annotators manage between 500 and 1,000 tokens per hour. This also highlights the varying demands posed by individual texts, as well as how some (experienced) annotators can increase their pace with 'easier' texts. Compared to the non-parsed version, pre-parsing yields an efficiency gain of at least 25–50%, raising productivity from about 350–650 tokens per hour to 500–1000 tokens per hour. This effect is even more pronounced with preparatory PoS tagging. Experienced annotators review pre-tagged data at a rate of approximately 2,000 tokens per hour, compared to 1,000 tokens per hour (values from the project's initial phase) for non-tagged versions, emphasizing the impact of typing/writing as a time factor. Manual analysis is generally influenced not only by the complexity of the annotation scheme and new texts that may still vary syntactically but also by (semi-automated) routines that can accelerate the process while still requiring oversight. Both the interaction between human and machine-based steps in the workflow, as well as each step individually (see the following Sections 2.2.3.1 and 2.2.3.2), must be evaluated and adjusted accordingly.

The ideal annotation tool can also be considered as part of the balance between machine-assisted and manual workflows. An ideal tool should present new suggestions from the parser or tagger after each new human annotation step. Fine-tuning of the parser model would also have to take place with the annotation created for each text, because texts that are conceptually oral, i.e., less standards-oriented, frequently exhibit new linguistic peculiarities.

### 2.2.3.1 Parser und Tagger

I begin this section by establishing a baseline for parser accuracy – first in general, then in applications comparable to GiesKaNe. Since relevant taggers were discussed in the derivation of HiTS, I refer you to Section 2.1.4. Although GiesKaNe, as presented in Section 1, follows a constituent structure in which syntactic functions are expressed via the edges, a dependency parser is used. If one takes into account the granularity of the annotation scheme or the number of syntactic functions (especially in the sentence) including the semantic roles, the main subject of the analysis is not form-constituents themselves, but the functions of their constituents. The choice of a constituent structure tree follows an established tradition for German treebanks (see Section 1) and does not involve a decision against syntactic relations as these are represented. Furthermore, the application to historical German is based on the idea that the dependency parser "performs better because it can handle long-distance relationships and coordination better" (Kübler & Prokić, 2006). However, Fraser et al. (2013) point out that it is "a widely held belief that dependency structures are better suited to represent syntactic analyzes for morphologically rich languages because they allow non-projective structures (the equivalent of discontinuous constituents in constituency parsing)". "However, this does not mean that dependency parsers function better here (cf. Tsarfaty et al., 2010). Above all, many dependency parsers would only generate non-projective structures from purely projective dependency structures when needed." Kübler's study (2008), which concluded that "higher accuracies for detecting grammatical functions with dependency parsers than with constituent parsers" can be achieved, is criticized by Fraser et al. (2013) due to the evaluation measure used (cf. Tsarfaty et al., 2012), which subsequently makes comparisons in different contexts appear questionable. This in turn becomes clear for GiesKaNe when transferring it to the spreadsheet format (see Section 2.2.4, Fig. 9).

Although there are central approaches to constituent structure parsing (e.g., Kitaev & Klein, 2018: F = 95.1; Tian et al., 2020: F = 96.4; Zhang et al., 2020: F = 95.69) within the value range subsequently determined for dependency parsing, a distinction must be made when parsing discontinuous structures, which would apply in this context. These lag significantly behind the continuous parsing approaches, but are getting closer. Fernández-González and Gómez-Rodríguez (2022: 13) give a very detailed overview of central approaches from 2015 to 2021 based on NEGRA and TIGER corpora, achieving a best F-score of 91.0, with a F-score of 76.6 measured exclusively on discontinuous constituents. For comparison, and to contextualize other values: When evaluated based on the Penn Treebank, the approach achieves an F-score of 95.23. As Fernández-González and Gómez-Rodríguez show (2022; cf. Fernández-González and Martins, 2015; but also Maier, 2015; Hall and Nivre, 2008), dependency-parsed structures or the methods used to create them make good use of the reconstruction or generation of constituent structures or they form the basis for a conversion (Bosco, 2007; Candito et al., 2010, Lee & Wang, 2016).[93] This conversion is simply rule-based in GiesKaNe and has not yet been adapted to the latest standards.[94] As Fernández-González and Gómez-Rodríguez (2022: 12) make clear, annotation schemes must be carefully adjusted to maintain accuracy. Fraser et al. (2013: 60) point out that constituent structures contain information that goes beyond dependency structures. In my opinion, such an effect should only become relevant in a high range of transmission accuracy.

Regarding dependency parsing: As already mentioned with PoS tagging, English, the Penn Treebank, the Wall Street Journal and the Stanford Dependencies Schema (de Marneffe & Manning, 2008) are the standard for evaluating the parsing methods themselves. The values for contributions between 2015 and 2022 are correspondingly high: UAS = 93.01 to 97.42, LAS = 91.0 to 96.26, PoS ≥ 97.3 (Weiss et al., 2015; Kiperwasser and Goldberg, 2016; Andor et al., 2016; Dozat & Manning, 2017; Clark et al., 2018; Fernández-González & Gómez-Rodríguez, 2019; Mrini et al., 2019, Tian et al., 2022).[95] The results of the CoNLL 2017 and 2018 Shared Task (Zeman et al., 2017 and 2018) based on the Universal Dependencies are correspondingly difficult to compare. Here the best approaches for the 'large' corpora achieved an LAS of 81.77 and 84.37, respectively. For the CoNLL 2008 Shared Task 'Joint Parsing of Syntactic and Semantic Dependencies', a best F-score of 85.95 was already achieved for the Wall Street Journal and 75.95 for the Brown corpus. For syntactic relations alone, the best F-scores were 90.13 and 82.81 (Surdeanu et al., 2008). So finding a baseline is difficult. Then there is the question of how the results can be related to German or even a historical language level. Rehbein and van Genabith (2007) respond to Kübler et al. (2006), who address the question of whether German is harder to parse than English. As a result, they consider the question to be still open and show that comparisons between

> "PARSEVAL-based parsing results for a parser trained on the TüBa-D/Z or TIGER to results achieved by a parser trained on the English Penn-II treebank [ …] does not provide conclusive evidence about the parseability of a particular language, because the results show a bias introduced by the combined effect of annotation scheme and evaluation metric."

Kübler et al. (2008), who also emphasize the influence of the syntactic annotation scheme on parser performance and evaluation, transfer 2000 sentences from TIGER and TüBa-D/Z into a dependency structure and achieve this with BitPar, LoPar and the Stanford parser for TIGER (LAS = 78.8 to 81.6, UAS = 83.0 to 85.6) and TüBa-D/Z (LAS = 71.3 to 75.9, UAS = 81.7 to 86.8). Eckhoff and Berdičevskis (2016) present an UAS of 0.845 and a LAS of 0.779 for Old East Slavic and the MaltParser, Pedrazzini (2020) for Old Church Slavonic an UAS of 83.79 and a LAS of 78.43 and for Old East Slavic an UAS of 85.7 and a LAS of 80.16. Nie et al. (2023) achieve an F-score of 67.3 for constituent structures in zero-shot parsing for Middle High German texts.[96] Sapp et al. (2023) make an interesting observation, which will also be included with a view to the distinction between conceptual orality and distance (Emmrich, 2024 and Section 2.3). When comparing the Berkeley Neural Parser (Kitaev et al., 2019) and the SuPar Neural CRF Parser (Zhang et al., 2020), they find that both perform equally for the text 'Neues Buch' (F-Score 42.38 and 43.12), while for the text 'Fierrabras' the BNP performs significantly better (F-Score 45.29 to 36.15). They conclude that 'Fierrabras' is temporally and dialectally closer to current Standard German and can therefore benefit more from word embeddings trained on modern texts. While I think the line of argument as mentioned is understandable, in the sense of the argumentation of Sapp et al. (2023) better values could actually be expected. For SuPar and 'Fierrabras' we see a fundamental decline that needs to be explained. With a relatively small test set ('Neues Buch': chronicle of the city of Cologne, 1360; 'Fierrabras': fiction, 1533; 'Historia': travel narrative, 1557) and a training on the Corpus of Historical Low German (Booth et al., 2020), Sapp et al. (2023) then show the different performance with F-scores of 35.45, 48.46 and 59.21. This also fits with the more extensive, early study by Versley (2005), who varies different methods, grammar models and text types and illustrates the influence on the results. This shows that the values vary significantly under very different aspects: parser method, amount of training data, variation of the texts in the data set, annotation scheme, language. And the parsing of the Wall Street Journal serves as a basis for comparison of the parsing methodology, but does not define the baseline for practical application.

---

[93] Fraser et al. (2013) are talking about machine translation systems that transfer dependency structures into constituents.
[94] An evaluation of possible advanced transmission methods is still pending for GiesKaNe.
[95] Sebastian Ruder maintains an extensive list: https://github.com/sebastianruder/NLP-progress/blob/master/english/dependency_parsing.md
[96] It is about a cross-lingual transfer technique that minimizes the necessary amount of training data in the target language. The approach seems particularly interesting in light of the derivation of an annotation scheme for further studies similar to that used for PoS tagging.

For the GiesKaNe corpus, we use SpaCy (3.0) and train the dependency parser[97] based on a Transformer model.[98] The BERT model – RoBERTa (Liu et al., 2019) – achieves an UAS of 95.1, a LAS of 93.7 and a $PoS_{accuracy}$ of 97.8 when parsing the Penn Treebank (English) compared to the optimized models. For GiesKaNe v.03 (as a dependency treebank) the following values are achieved with an 80:20 split for training and test data: UAS = 0.88, LAS = 0.77, $PoS_{Acc}$ = 0.94. The above-average demanding PoS tagset (auxiliary verb for passive, copula verb, semi-modal verb, aci-verb, functional verb) was already addressed in Section 2.1.3. In this respect, the value appears reasonably good. The decreasing value range during parsing is to be expected for these texts. However, it could also go hand in hand with the granularity of the dependency parser labels, which contain a lot of semantic information as partially discussed in Section 2.1.3. As mentioned, a loss of accuracy is to be expected when converting to a constituent structure, but this has not yet been evaluated.[99]

By directly transferring the parsed dependency structure to a spreadsheet (constituency structure), the differences in formats become clear. Here each token $_{tn}$ corresponds to a row and each column represents a syntactic depth $d_m$ for $t_n$, which results from the syntactic functions $d_m$ for $t_n$ and the form constituent that realizes it ($d_{m+1}$ for $t_n$) (see Fig. 9.2). This is not the case with dependency structures – if you consider the CoNLL format (e.g., Nivre et al. 2016; for CoNLL-U) with its single ID relations – for example. However, what can be compensated for in the calculation (e.g., tree edit distance) is a corresponding additional effort when focusing on human-machine integration (Section 2.2.3).

As mentioned, experience reports, studies, and a review of the literature indicate that parser accuracy values, while representing a controllable starting point, are in themselves less meaningful since each individual text introduces unique challenges. Consequently, an ideal tool would need to make real-time suggestions, and the model would require fine-tuning to the specific characteristics of the current text. Parsers are often developed not solely for supporting manual annotation but also with a focus on optimizing methodology or meeting the specific requirements of a given dataset. However, manual annotation could benefit even further if parsers were adapted to interact more effectively with human annotators in such a way.

### 2.2.3.2 Inter Annotator Agreement

As Arstein (2017: 297) summarizes, the Inter-Annotator Agreement (IAA) serves the goal of validating and improving the annotation scheme and the corresponding guidelines in the manual. It is intended to reveal ambiguities as problem areas. However, the annotation process in a broader sense is also put to the test. With this objective in mind, I will discuss the benchmark for comparison and the values achieved in comparable projects. As noted by Dipper et al. (2024) as part of their evaluation of the Universal Dependencies Extension for Modern and Historical German, chance-corrected evaluation measures such as Cohen's κ (Cohen, 1960) are generally used for flat annotations.[100] According to my research, this does not exist for dependency relations, aside from the contributions discussed below. Instead, alongside the F-score[101], there are LAS and UAS[102], both of which are not chance-corrected. Brants and Hanse (2002) also use an F-score for the TIGER corpus and evaluate the annotation scheme for individual labels, but finally present an overall F-score (93.89). Dipper et al. (2024), like us, take up the ideas of Skjærholt (2014), who uses the tree edit distance (Zhang & Shasham, 1989) based on the Damerau/Levenshtein distance (Damerau, 1964; Levenshtein, 1966) to calculate a value in the sense of Krippendorf's α (Krippendorff, 1970/2004). For the Middle High German texts, Dipper et al. (2024) presented an α value of 0.85 compared to UAS = 0.89 and LAS = 0.79 in a comparative evaluation of the extension of the UD annotation scheme. What is interesting is how they assess the values. Dipper et al. (2024: 17107) note: "In general, chance-corrected scores of 0.61–0.80 are often considered 'substantial' and scores of > 0.81 'almost perfect agreement' (Landis and Koch, 1977)." In contrast, Arstein and Poesio (2008: 576) say:

> "Researchers have attempted to achieve a value […] above the 0.8 threshold, or, failing that, the 0.67 level allowing for ‚tentative conclusions.' However, the description of the 0.67 boundary in Krippendorff (1980) was actually ‚highly tentative and cautious,' and in later work Krippendorff clearly considers 0.8 the absolute minimum value […] Recent content analysis practice seems to have settled for even more stringent requirements […]."

---

[97] https://spacy.io/api/dependencyparser

[98] de_dep_news_trf (3.7.2), https://spacy.io/models/de

[99] In such a comparison, confounding factors and the measure or method of comparison would have to be taken into account in the sense of the differences mentioned (cf. Rehbein & van Genabith (2007); Kübler, 2008; Tsarfaty et al., 2012; Fraser et al. (2013)).

[100] A relevant example is provided by Koleva et al. (2017), who evaluated the HiNTS tagset as part of their presentation of a PoS tagger for Middle Low German and reported kappa values of around 0.91 to 0.92.

[101] A relevant example of this is the analysis of the IAA in Rehbein et al. (2012) as part of the investigation of the effects of pre-annotations on manual annotation.

[102] Biagetti et al. (2021) provide these values for the Treebank of Vedic Sanskrit (UAA = 76.0% and LAA 0 63.8).

Neuendorf's (2002) overview is echoed by Arstein and Poesio (2008) to the effect that "reliability coefficients of .90 or greater would be acceptable to all, .80 or greater would be acceptable in most situations, and below that, there exists great disagreement. Lynn (2016: 52 ff.) uses this α value to improve the annotation scheme directly with 2 IAA studies and reports an improvement from 0.79 to 0.85 between IAA1 (LAS = 0.74; UAS = 0, 85) and IAA2 (LAS =0.79; UAS = 0.88). A detailed IAA study for GiesKaNe was conducted during an early project phase to evaluate the annotation scheme within the overall workflow. The evaluation of the treebank was supplemented by an analysis of PoS tagging and central span annotations (e.g., sentence boundaries) at the macro level. For these flat annotations in the sense of sentence boundary segmentation, Lopatková et al. (2012) present a comparison value, which, however, only indicates a percentage agreement – in the range of around 94% when determining the segments.

The IAA study on GiesKaNe is based on 5 texts, with 1,000 tokens from each text analyzed across these levels. Two annotators were always compared according to the actual work steps for the 3 areas of segmentation at the macro level (sentences, non-sentences, cohesion units), the syntactic functions ('Satzglieder') including semantic roles and word groups (meso level) as well as parts of speech (micro level). In total, 15 partial analyzes were carried out. The tree structures in the core of the GiesKaNe treebank, which are more difficult to evaluate, were done in 3 ways in order to gain more insights into the annotation process, which enables improvements, in addition to comparing them with other projects. Krippendorf's alpha was recorded in two ways: via form-function label combinations as nodes of a graph and via the individual cells in the spreadsheet according to the mapping of tokens ($t_n$) and their depth ($d_m$) – as shown at the end of Section 2.2.3.1 or in Fig. 9.2. The latter measure – practically based on our annotation process – includes the work effort for evaluating each cell of the spreadsheet. In addition, the tree edit distance was also considered. The graph-based approach does more justice to the idea that was introduced in Section 2.2.3.1: A whole lot of misplaced cells can be adjusted by inserting a single level. In this respect, it is not about absolute matches between cells in a table, but relative agreement of structures. However, as mentioned, the number of corrections is also a relevant factor given the human workload.

| | Graph | cells of the spreadsheet | |
| --- | --- | --- | --- |
| | | syntactic function | phrase |
| **Becher** | 0.89 | 0.95 | 0.89 |
| **Bauernleben** | 0.92 | 0.90 | 0.93 |
| **Koralek** | 0.94 | 0.93 | 0.95 |
| **Nehrlich** | 0.92 | 0.93 | 0.94 |
| **Ranke** | 0.84 | 0.86 | 0.91 |

Fig. 8, Krippendorf's alpha, calculated via graph and spreadsheet

The values in Fig. 8 are initially unproblematic given the values in comparable projects, the values in machine analysis and the granularity of the annotation scheme.[103] However, the pure value can only play a secondary role. Comparisons with other projects and value ranges ultimately provide no insight into optimizing the annotation scheme, and the differences between the texts in Fig. 8 offer no basis for drawing meaningful conclusions. The tree edit distance (TED) is more informative here. The mean TED for Becher is 4.45 (std = 7.28) and for Ranke it is 5.4 (std = 7.13), while for Bauernleben, Koralek and Nehrlich it is 2.02, 1.55 and 1.86. As already mentioned when discussing the annotation scheme in comparison to standards in Section 2.1.3 or the normalization in Section 2.2.1, the distinction between conceptual orality and literacy ( Section 2.3) could be relevant here: Ranke and Becher are conceptually literary texts from the fields of history and economics. However, the average token length per sentence here is also around 18 and 20. That of the other texts is between 9 and 10. The syntactic depth (calculated based on the number of tokens $t_{n+1}$ and their respective syntactic depth $d_{m+1}$) is 3 or 4 times higher in both conceptually literal texts. In the multiple linear regression, a model with all 3 independent variables only shows a significant contribution from tokens and their depth (with p < 0.001 each): $R^2$ adjusted: 0.649, p < 0.001. From a linguistic-theoretical perspective, it is certainly logical that there is not necessarily a dependency relationship per se between sentence length in tokens and syntactic depth. Multicollinearity does not appear to be an issue here (VIF = 5.86). The tokens also show a negative regression coefficient of -1.15. The Drop1 test, which evaluates the impact of removing each predictor,  highlights the relevance of both values, but also indicates a more pronounced effect of syntactic depth (F = 477.9 vs. 191.48, both p < 0.001). Overall, this seems to indicate that generally longer, but syntactically less deep sentences lead to less TED and greater syntactic depth leads to more TED. Biber and Gray (2010) compare the clausal discourse style and the phrasal discourse style. This corresponds to the fact that the proportion of TED changes 'Insert' and 'Remove' is significantly higher than 'Update' for the two distance texts (conceptual literacy, see Section 2.3) with greater syntactic depth. The annotators' attention does not have to be explicitly drawn to the complexity of the structures, but rules for dealing with ambiguities must be emphasized. The consequences of errors in high nodes when correcting the parser can also be pointed out once again and you should plan more time overall. The TED

---

[103] With upcoming studies, we have to ensure – also with regard to final control routines for the corpus – that not only annotators agree, but that they as a group (still) agree with the annotation manual. You would now have to carry out an analysis by a third party based solely on the annotation manual and use this as a basis for comparison.

Updates are also an interesting point in the IAA study: Recognizable partial problems in this study were different semantic roles with the same syntactic function (21.3%), different syntactic functions (11.4%) and subtypes of adverbials (6.6 %). At the phrase level, coordination emerges as a problem area (19.2%). Here too, one can see that similar to a parser, agreement decreases when the analyzes move away from the text surface/form level and open up space for interpretation in semantic-logical dimensions. The annotation scheme could be refined early on in a rule-based manner and with lists of case groups. In addition, a high alpha value of 0.963 for the macro elements and 0.978 for the subordinate clause can be determined, which also appears to be unproblematic. When analyzing the heterogeneous group of 'other structures', the values then drop significantly. It involves the annotation of subordinate clauses that refer to several main clauses ('Mehrfachgeltung'), coordination ellipses or speech indicators. However, the alpha score is not an appropriate measure here either. Because at its core, this is about recall. Annotations that only need to be done occasionally require a different form of attention/different routines to be anchored in the workflow. For example, quickly reviewing a passage of text from this perspective. For the parts of speech, the alpha value is 0.958. So roughly in the area of deriving HiTS from existing annotation (Section 2.1.4) and slightly above the parser value. Corresponding problems were addressed there. It should be emphasized once again that the PoS annotation is only checked one-sidedly at the end of the process, which is why they require special attention. Problems here arise from the granularity of the tagset or from the interpretation detached from the surface structure (verb subclasses for passives, semi-models and aci-verbs or additive annotation (see Section 2.1.3 and 2.2.3.1). One strategy is to use differently trained models when tagging, comparing their results and highlighting the differences in color. Zinsmeister (2015: 96) already suggests the use of multiple PoS taggers simultaneously as a strategy to compensate for the 'weaknesses' of the individual taggers or – with reference to Loftsson (2006) or Rehbein (2014) – to identify errors through differing analyses at certain tokens. In principle, it is now possible to combine vector-based models with a CRF using selected features or to establish two focal points with distinct feature sets. Alternatively, it would also be conceivable to have a single model output narrow decisions along with multiple possible tags. In addition to our spacy model, we also use a CRF with morpho-syntactic features that was initially employed during the early phase of the GiesKaNe project.[104] Example 8 provides a final insight into the application.

**Example 8** (Neitzschitz, 1666): […] bin also neben ihm / mich <u>um</u> zu sehen / durch die Stadt geritten […][105]

Engl.:                                        […] so I rode alongside him / looking around / through the city […]

If you read Example 8 (as a human), the very likely assumption of (two) constituents between the finite verb *bin* and the pronoun *mich* makes the probability of an interpretation of *um* as a subjunctor (without a comma) linked to *mich* approach zero, which means *um + zu + sehen* as the to-infinitive of the particle verb *umsehen* becomes highly likely. The problem is of course 'homemade' on the basis of the modified tokenization (*umzusehen* becomes *um + zu + sehen*). However, the need for a finer tokenization with a uniform reference to the text surface was also addressed. The two tagger models do not match (in the sense of taking both options into account) and this can be automatically highlighted in color in the spreadsheet.

## 2.2.4 Spreadsheet as an annotation tool

The work with a simple spreadsheet arose from the fact that Annotate (Brandts & Skut, 1998) as the standard tool for the annotation of constituent structures (in German studies) has not been maintained for a long time (cf. Dipper and Kübler, 2017). New projects such as Hexatomic (Druskat et al., 2014) were unable to close this (large) gap – at least at the beginning of the project in 2016. In addition, when developing earlier annotation tools, a generic orientation was rarely considered, so that the grammatical model (here: the GTA) could not necessarily have been made the starting point of the work. Basically, through the Universal Dependencies (UD) projects, the CoNLL formats or NLP libraries like Spacy, you can always find good starting points for working with dependency structures – such as Arborator (Gerdes, 2013)[106]. For constituent structures – especially against the backdrop of discontinuous structures – things do not look quite so good. Tools such as WebAnno (Yimam et al., 2013) and INCEpTION (Klie et al., 2018) are based on Brat (Stenetorp et al., 2012) and are suited for dependency structures. Another approach could be LightTag (Perry, 2021) or Prodigy.[107] Spreadsheets have also

---

[104] The approach could be refined as mentioned.
[105] Neitzschitz, Georg Christoph von: Sieben-Jährige und gefährliche WeltBeschauung Durch die vornehmsten Drey Theil der Welt Europa/ Asia und Africa. Bautzen, 1666. , p. 154 f. In: Deutsches Textarchiv.
https://www.deutschestextarchiv.de/neitschitz_reise_1666/160
[106] https://arborator.ilpga.fr/
[107] In addition to tools like Prodigy (https://prodi.gy) from the Spacy Framework, there is an overall trend towards more and more tools from an area at the interface between science and business. From a scientific perspective, it is still surprising that

been used for annotations, but more often for flat ones. Odebrecht et al. (2017) use a spreadsheet as part of the RIDGES Herbology Corpus. The GUM corpus is also partially annotated using GitDOX (Zhang & Zeldes, 2017) based on a spreadsheet editor. The Praaline tool/system for spoken language corpora includes an editor in spreadsheet form for annotations. The practice is also supported by De Cock et al. (2024: 86), who address easy availability.

Overall, however, the use is rather selective. A spreadsheet basically has a few advantages (I repeat briefly from Section 2.2.2): Spreadsheets can be used on any PC, spreadsheets are stable, MS Excel offers interfaces to extensive Python APIs, is suitable for collaborative work in the cloud, offers freedom for any additions (including comments), spreadsheets offer graphical support and the possibility to automate work steps. Nevertheless, the desirability of a tool in the tradition of Annotate—featuring parser suggestions after each manual annotation and integrating ideas for collaborative work, as seen in the WebAnno tool group—is not diminished by the use of what is already available.

Fig. 9 shows the annotation process: Fig. 9.1 is the dependency-parsed structure, which is transferred to a spreadsheet based on rules. In Fig. 9.2, the maximum span of the sentence is at the beginning (far left) – in contrast to the minimum sentence node at the top of a (ordinary) tree graph as in Fig. 9.3. Functions and forms then alternate in the columns (Fig. 9.2). Discontinuous nodes are marked with angle brackets. Here, we use a basic rule of the GTA grammar model (Ágel, 2017): Since each syntactic function can only occur once within a form constituent, the function labels can be used as IDs that uniquely identify a cell or connected cell. The conversion from spreadsheet to XML format uses the property of spreadsheets that each cell can basically be identified by numbers in columns and rows, so that you can determine which cell or merged cell (column right) belongs to which level (column left). This allows the constituent structure trees to be easily generated. An API like openpyxl[108] is used for this. The target format is the XML standoff format PAULA (Dipper, 2005). This is then transferred using the graph-based converter Pepper (Zipser & Romary, 2010) into the format required for ANNIS (Krause & Zeldes, 2016). Part of the tree is shown in Fig. 9.3.[109]

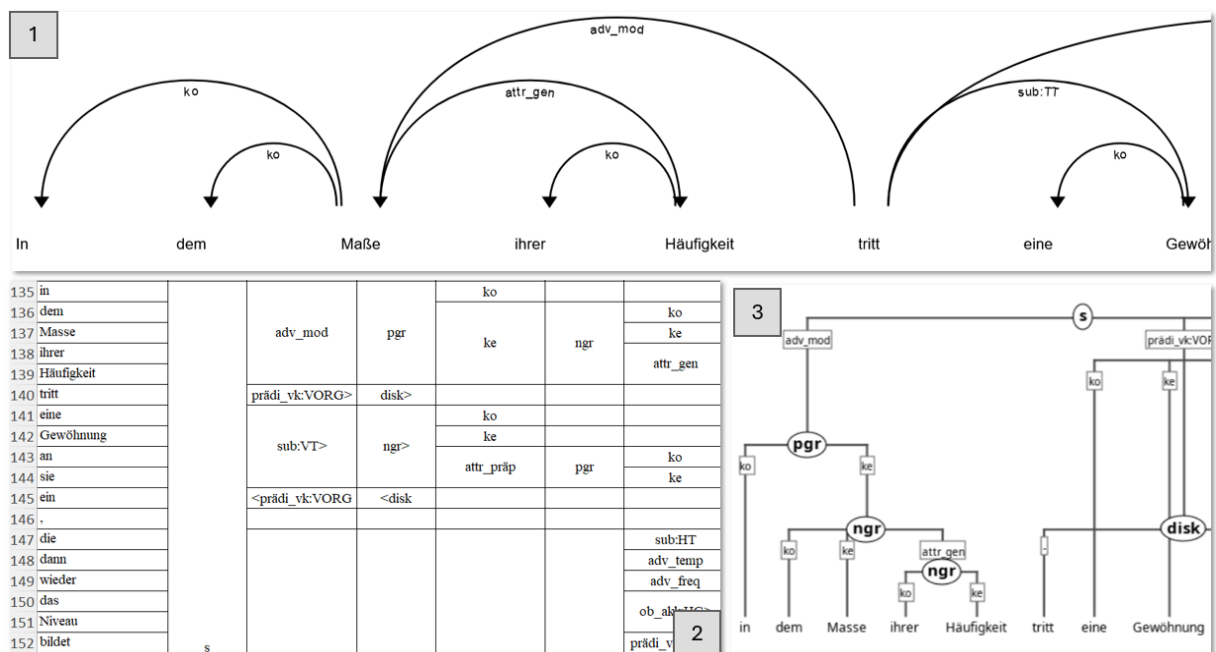

Fig. 9, dependency parser analysis (1), spreadsheet representation (2) and constituent structure of GiesKaNe in ANNIS (3)

With these concluding remarks on the spreadsheet as a central tool in our workflow, the established workflow—primarily examined from the perspective of the interaction between human and machine-assisted work steps—can now be finalized. In Section 2.3, I conclude with the machine-based determination of conceptual orality and literacy, a topic that has emerged from more recent developments.

---

there is not a single tool that has been maintained in long term. Ultimately, annotations of texts can be built on a few basic structures.

[108] https://openpyxl.readthedocs.io/en/stable/
[109] The whole tree in GiesKaNe v03: https://annis.germanistik.uni-giessen.de/?id=d34a0b20-5aec-406f-86df-8bb3ed8177d5#_q=bm9ybSA9Ik1hw59Ii4iaWhyZXIi&ql=aql&_c=R2llc0thTmVfdjAuMw&cl=10&cr=40&s=0&l=10&m=0

## 2.3 Compilation of texts: Automatic determination of conceptual orality and literacy and the representativeness of text segments

The text selection was already mentioned at the beginning of this article. Basically, it is about representativeness, which, as Egbert et al. (2022: 30 ff.) show that it can be viewed from several perspectives. Among other things, the 'absence of selective force' needs to be discussed here, because 'random sampling' would not be possible and Egbert et al. (2022: 32) contrast this view with statements by Johansson et al. (1978): "The true 'representativeness' of the present corpus arises from the deliberate attempt to include relevant categories and subcategories of texts rather than from blind statistical choice." For GiesKaNe, Fig. 1 addresses a division into three centuries (17th, 18th, 19th century) and 4 text domains/text types (everyday texts, science, utility literature, fiction).[110] In principle, however, additional dimensions must be taken into account here. Biber et al. (1998: 252; cf. Meurman-Solin, 2001) note three points:

> „One major issue in designing a multi-purpose [diachronic] corpus is deciding what registers to include. Three concerns arise here: 1. to choose registers that represent a wide range of the kinds of writing found in a given historical period; 2. to include 'speechbased' registers (e.g. drama or court testimony), to provide some idea of the characteristics of spoken language in the historical period (since there are no recordings of actual speech from earlier periods); and 3. to choose registers that have a continuous history across periods."

Meurman-Solin (2001: 9) again refers to Biber et al. (1998) point out that genre categories exhibit internal heterogeneity: "For example, medical research articles in the early 1700s were typically case studies written as personal letters to the editor of a journal; this is in contrast to the dense experimental journal articles typical of the twentieth century." (Biber et al, 1998: 252) For the Early Modern English part of the Helsiki corpus she quotes Nevalainen and Raumolin-Brunberg (1989: 95): „[T]ext classification involves a scalar notion of genrehood with some nearly invariant generic structures at one end (letters, statutes) and some quite loosely structured ones at the other (book-length treatises on different topics)." Whitt (2018: 4) also points out corresponding connections:

> „Another challenge (and benefit) to users and compilers of diachronic corpora is the ability to represent language change both 'from above' and 'from below' (Labov 1994: 78; cf. Rissanen 2009: 57); […] Closely related to this are the notions of immediacy and distance in language (Koch & Oesterreicher 1985, 2007, 2012; see also Elspaß 2014): some genres are more representative of face-to-face, spoken, interpersonal ("immediate") communication while other genres are of a more impersonal, written, and formal ("distant") nature, and a continuum exists between these two poles. Biber's (1986, 1988) multi-dimensional approach to variation makes a similar contrast."

Overall, it can be said that, especially in the area of a broad register concept, there are dimensions that must be taken into account or controlled that cannot be handled via simple domain or text type assignments. The implementation is not that easy or remains quite abstract in most corpora. Farasyn et al. (2018: 286) address this topic with a concise remark in the context of text selection for the parsed corpus of Middle Low German: „In these scribal languages, the writers/scribes did not try to represent the local dialect, and the difference between spoken and written language might well have been considerable (Fedders 1988)." The approach of the Indian Parsed Corpus of (Historical) High German (Sapp et al., 2024) is more practically oriented: "In order to capture features of less formal, even stigmatized language (Schäfer, 2023), we try to include 1-2 dramas per time bin." GiesKaNe builds on the KAJUK (The Kassel Corpus of Clause Linking), for which Hennig (2013) also describes the compilation with reference to the concept of conceptual orality and literacy ('Nähe und Distanz', Koch & Oesterreicher, 1985). Hennig and Ågel (2006) „developed the Koch/Oesterreicher model further by focusing on compiling grammatical features typical of immediacy (conceptual orality), and relating them to pragmatic conditions, such as whether the roles of the participants as producers or recipients are fixed or flexible, whether the producers can take their time to plan their utterances or are forced to speak by the presence of the recipients, and so on (Ågel & Hennig, 2006). Niehaus and Elspaß (2018) also use the concept in their compilation of the 'NiCe German Corpus'. In both cases, the original classification as immediacy or distance text is not based on any prior analysis.

Rather, the analysis of characteristics takes place retrospectively and is evaluated based on the previous classification. This is based on an assessment of external communication conditions, while the interaction between internal and external conditions is at the center of Koch and Oesterreicher's model. This practice is logical, but can be linked to the question of how to enter into a chicken-and-egg problem cycle. Here we try to get started using Emmrich's (2024) approach through automatic analysis and thus determine in advance the relative value of a text on a scale of conceptual orality and literacy (COL-scale). One can compare it with Biber's multi-dimensional analysis or Dimension 1 (1988). However, Biber uses Common Factor Analysis (CFA) to explore different dimensions, while PCA is used here to create a quasi-tailored artificial 'dimension' in a controlled manner in order to be able to explain other conditions. Starting points for mapping external communication conditions through internal linguistic features are provided by Ägel and Hennig (2006), whose work serves as a starting point for the features of conceptual orality. A binary classification of the texts by experts is used to select the characteristics. In this way, the best correlating binary predictors such as sentence length, interjections and speaker-listener pronouns, but also syntactic complexity and information density (overall 9 features) are determined and ultimately reduced to a scale via PCA (Fig. 10).

---

[110] Other projects such as the A Penn-style Treebank of Middle Low German also incorporate dialect spaces. Booth et al. (2020) point out that this cannot always succeed.

| Century | Text type | Value | Name |
|---|---|---|---|
| 19 | Correspondence | -3.8 | N_briefwechsel |
| 18 | Autobiography | -3.1 | N_nehrlich |
| 17 | Autobiography | -2.9 | N_guentzer |
| 17 | Diary | -2.7 | N_soeldnerleben |
| 17 | Chronicle | -2.7 | N_bauernleben |
| 19 | Diary | -2.6 | N_koralek |
| 18 | Autobiography | -2.0 | N_braeker |
| 19 | Diary | -1.9 | N_zimmer |
| 18 | Autobiography | -1.8 | N_dietz |
| 17 | (Science) History | 0.5 | D_freyberger |
| 17 | (Use) Decency Lit. | 0.6 | D_nn-weltmann |
| 18 | (Science) Theology | 0.7 | D_mendelssohn |
| 17 | (Use) Society | 0.9 | D_birken |
| 18 | (Science) History | 0.9 | D_gessner |
| 19 | (Use) Travel literature | 1.3 | D_michelis |
| 19 | (Science) History | 1.5 | D_ranke |
| 18 | (Use) Decency Lit. | 1.5 | D_knigge |
| 19 | (Use) Society | 1.7 | D_anthus |
| 17 | (Science) Economics | 1.8 | D_becher |
| 17 | (Science) Philosophy | 1.8 | D_thomasius |
| 18 | (Use) Popular Science | 2.0 | D_harenberg |
| 18 | (Science) Travel Lit. | 2.1 | D_forster |
| 19 | (Science) Philosophy | 3.1 | D_nietzsche |
| 19 | (Science) Sociology | 3.4 | D_simmel |

Abb. 10, scale of conceptual orality and literacy according to Emmrich (2024) with text type and century and binary classification (N = c. orality, D = c. literacy)

In addition to the text selection in the corpus compilation, Emmrich (2024) and Emmrich et al. (2024) also showed other possible applications: application to large corpora such as the DTA for orientation and text selection, as an explanatory variable or as a way to control possible influencing factors on a dependent variable. The possibility of estimating the workload for individual texts in the workflow in advance has already been mentioned. What is particularly interesting here with respect to corpus compilation is how the corpus evolves with each new text. The scale via PCA in Emmrich (2024) is a relative scale in the sense of Koch and Oesterreicher's theory (1985) and changes in practical application with the texts included. In this respect, each new text can be checked for how it complements the corpus. As part of the corpus compilation, another important question can also be addressed in connection with the scale. GiesKaNe carries out extensive manual annotations and has to limit entire texts to text segments. So it is not just about selecting texts, but also about selecting text segments. According to Biber and Jones (2009: 1289)

> "The number of samples from a text also deserves attention, because the characteristics of a text can vary dramatically internally. A clear example of this is experimental research articles, where the introduction, methods, results, and discussion sections all have different patterns of language use. Thus, sampling that did not include all of these sections would misrepresent the language patterns found in research articles."

Fig. 11 shows the texts of the GiesKaNe corpus contained in the DTA corpus. The total texts were each divided into n segments of around 12,000 tokens as in the GiesKaNe corpus. Each segment of a text except the first and the last shares half of the tokens with an adjacent segment ($Seg_1$ = token 0 to 12,000, $Seg_2$ = token 6,000 to 18,000, $Seg_3$= token 12,000 to 24,000, ...) to avoid hard breaks. The box plots are sorted by the mean, while the black line indicates the median and the dot indicates the value of the corresponding text segment in GiesKaNe. The mean IQR for the texts is 0.51 (sd = 0.26) and varies between 1.0 for Birken and 0.1 for NN-Weltmann. A higher IQR seems to occur in the texts that tend to be conceptually oral.[111] If the sampling nature of text segments is taken into account, no major problems emerge. Nevertheless, it also becomes clear that special and mostly small corpora should not be viewed without reference to more extensive corpora.

---

[111] A visual inspection does not indicate that text length has a noticeable effect on the variation in values. However, the impact of short texts (cf. Liimatta, 2023) must also be examined in further development.

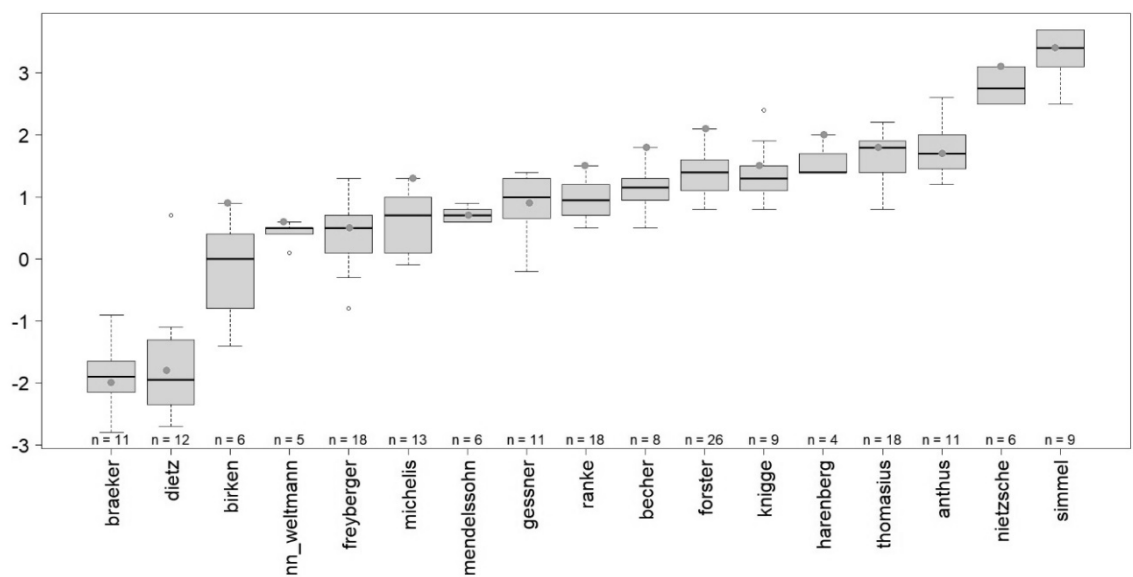

Fig. 11: COL-value of the text segment in GiesKaNe (dot) in relation to n segments of the same length of the entire text in the DTA corpus

As for the cases in which the GiesKaNe text value (dot) lies outside the IQR (Birken, Michelis, Ranke, Becher, Forster, Harenberg), this shows that automatic analyzes such as in this approach are essential for corpus compilation – not only in the dimension of 'Nähe und Distanz'. If you look at the IQR or median of the complete texts, the order of the texts on the COL-scale shifts slightly compared to that determined for the GiesKaNe texts. This does not affect the relationship between features and COL-value, but weakens the possibility of relating feature values and COL-value to the external conditions of communication. As I said: The text segment fully uses the metadata of the entire text, unless further differentiated. This in no way reduces the value of qualitative analyzes with a focus on text-functional or thematic aspects, but rather should be seen as a supplement. This last section can also highlight that, as the corpus grows, not only do more complex decisions need to be made in coherent analysis and expansions, but the corpus also becomes a valuable resource. As shown here, it can be used as training data or as a basis for comparison to support these tasks. With this, I conclude with a final look at GiesKaNe's workflow and this article.

## 3. The GiesKaNe corpus and its workflow: summary, conclusions, outlook

The starting points for the selection of topics for this article were the various dimensions within which GiesKaNe has to meet requirements that then shape the compilation process. Central factors include its size and the workload invested in its creation. Both aspects result in an orientation towards potential usage scenarios in the research community, extending beyond the internal research interest in syntax. As a historical corpus, GiesKaNe must connect historical language stages and create links to other historical corpora, but also to contemporary language. While these are characterized by large amounts of text and variation in types of text in written language (increasingly supplemented by spoken language), historical corpora face the challenge of depicting everyday language in the medium of writing, which, as recently shown, requires greater effort.

Fundamentally challenging is the preparation of the historical text's surface in terms of tokenization and normalization, as well as text segmentation. All these steps were machine-supported, referencing an extensive discourse, and they were intended to illustrate the importance of human-machine interaction in the workflows of extensive historical corpora. That is why this topic has been addressed here several times. GiesKaNe also had to be positioned between the comprehensive grammar model of the GTA, which is oriented towards practical text analysis, and possible de facto standards (historical and contemporary). In this context, an effort was made to integrate the advantages and disadvantages of such standards into the discussion of the grammar model.

Basically, interoperability as well as usability on the one hand and flexibility and originality on the other hand are opposed. This dichotomy underscores that annotation schemes of contemporary language treebanks cannot describe the historical syntax. Using the example of HiTS, it was shown how multiple interests can be addressed if a de facto standard in machine analysis is derived from verified annotations. In this context, the terms additive and consecutive annotation were used to problematize the idea that different annotation levels should not necessarily confirm each other but rather complement each other, although aspects of usability and interoperability must also be taken into account. Finally, with reference to the concept of 'immediacy and distance' ('Nähe und Distanz'), a possibility of automatic support for corpus compilation was emphasized.

Since the entire annotation process is effectively tied to a spreadsheet, this approach was used as an opportunity to demonstrate that not every project needs to invest in its own (short-lived) annotation tool. Instead, future investments should focus on more generic tools that can offer refined methods. For the parser, the emphasis was placed on interaction with human annotators. Here, the parser should also be given greater focus as part of the annotation tool. Regarding GiesKaNe, which is being developed at the project locations in Gießen and Kassel, the advantages of functions supporting collaborative work are clear. Above all, an annotation tool must be designed to be as generic and theory-neutral as the annotation levels in the formats and platforms themselves, in order to justify the significant development and especially maintenance efforts. The importance of resources for collaborative corpus compilation cannot be overstated when considering the future of GiesKaNe. Ideally, the corpus will expand in breadth but, more importantly, in depth through further collaboration, enabling the full potential of its complex annotations to be realized.